%% file: main.tex
\documentclass[10pt,twocolumn,letterpaper]{article}

\usepackage[pagenumbers]{cvpr} 


\usepackage{graphicx}
\usepackage{amsmath}
\usepackage{amssymb}
\usepackage{booktabs}
\usepackage{pifont}
\usepackage{multirow}
\usepackage{comment}
\usepackage{adjustbox}

\definecolor{cvprblue}{rgb}{0.21,0.49,0.74}
\usepackage[pagebackref,breaklinks,colorlinks,allcolors=cvprblue]{hyperref}

\def\paperID{} 
\def\confName{CVPR}
\def\confYear{1999}

\title{DarkIR: Robust Low-Light Image Restoration}

\author{Daniel Feijoo~$^{1}$, Juan C. Benito~$^{1}$, Alvaro Garcia~$^{1}$, Marcos V. Conde~$^{1,2}$ \\
{\normalsize $^1$~Cidaut AI, Spain}\\
{\normalsize $^2$~Computer Vision Lab, University of Würzburg}\\
{\tt\normalsize {\url{https://github.com/cidautai/DarkIR}}}
}

\begin{document}

\twocolumn[{%
\renewcommand\twocolumn[1][]{#1}%
\maketitle

\vspace{-7.5mm}
\begin{center}
    \setlength{\tabcolsep}{1pt}
    \begin{tabular}{c c c}
         \includegraphics[width=0.32\linewidth]{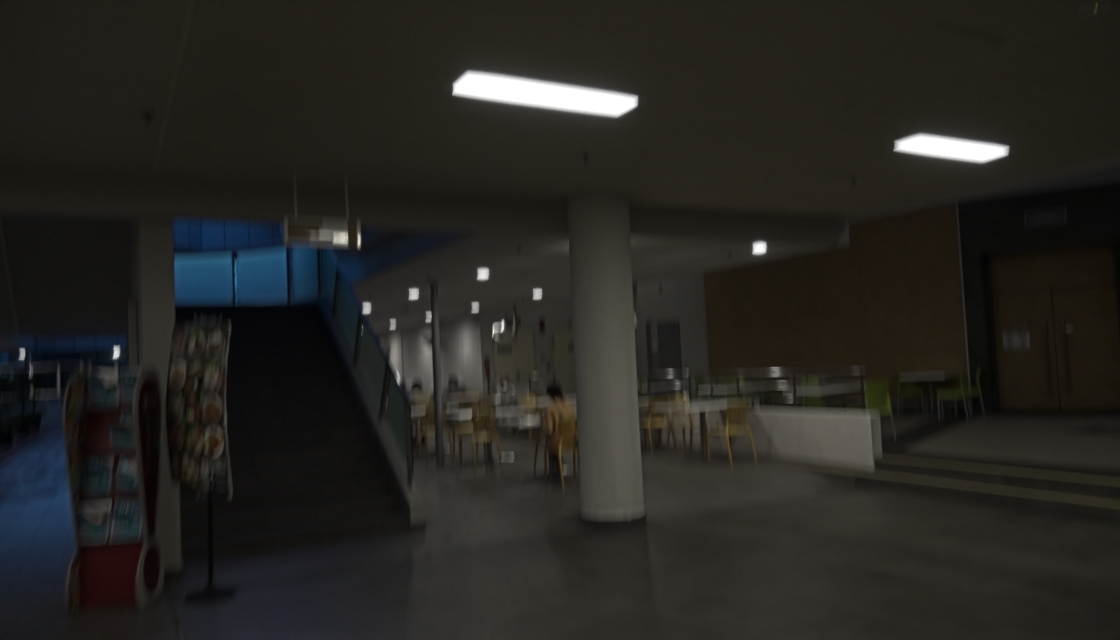} &
         \includegraphics[width=0.32\linewidth]{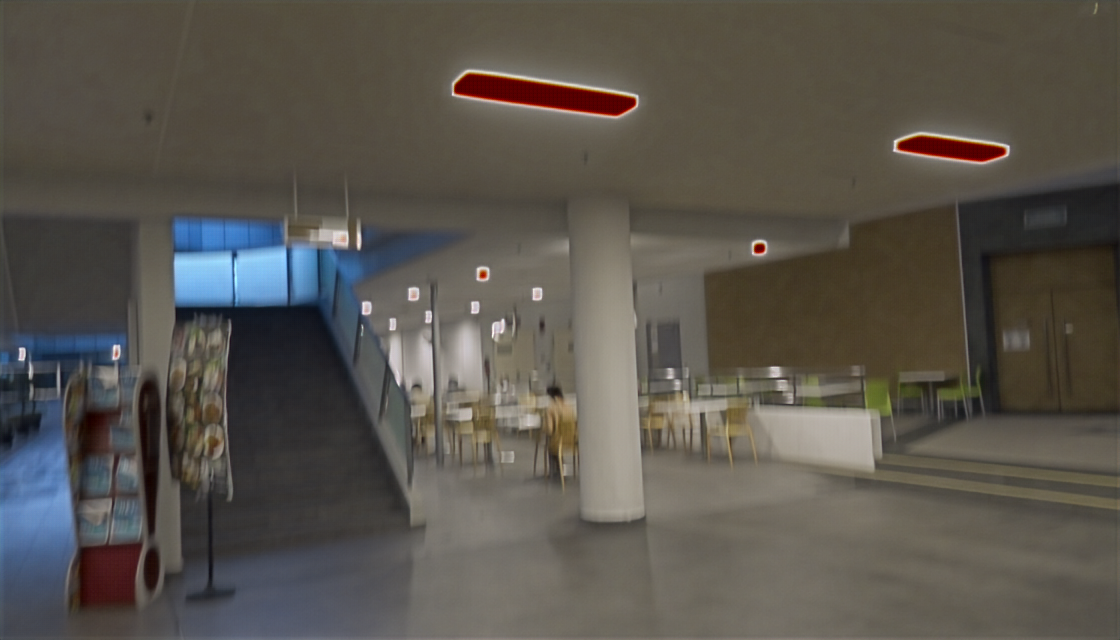} &

         \includegraphics[width=0.32\linewidth]{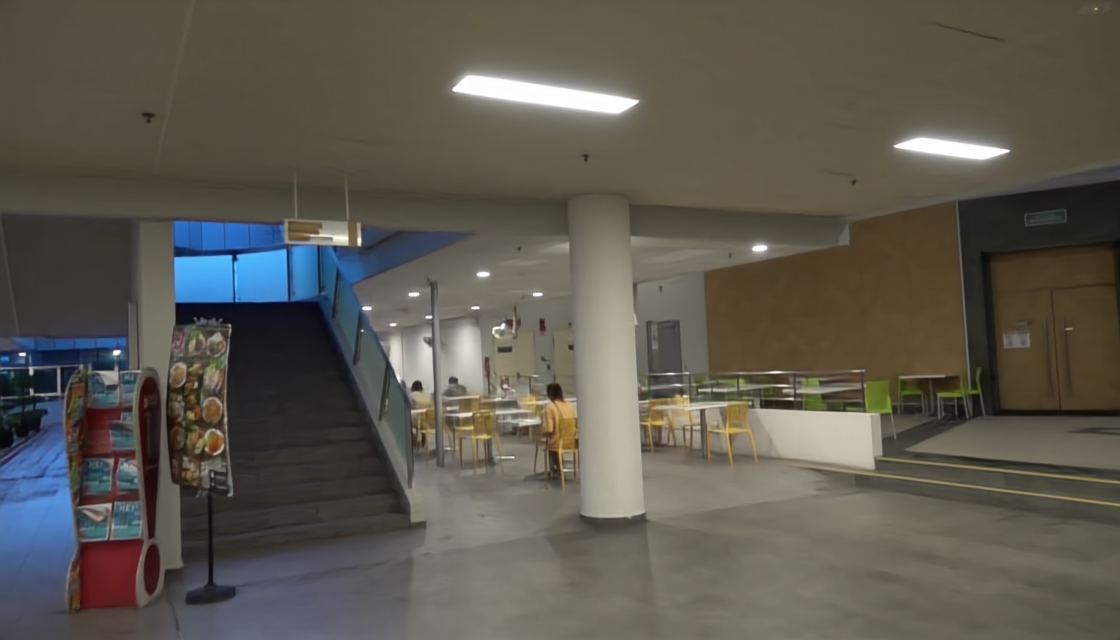} \\

         Low-light image with Blur & RetinexFormer~\cite{cai2023retinexformer} & \textbf{DarkIR} (Ours) \\
         
         \includegraphics[width=0.32\linewidth]{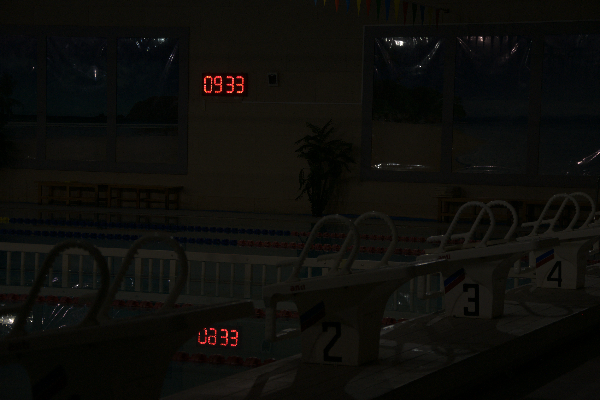} &
         \includegraphics[width=0.32\linewidth]{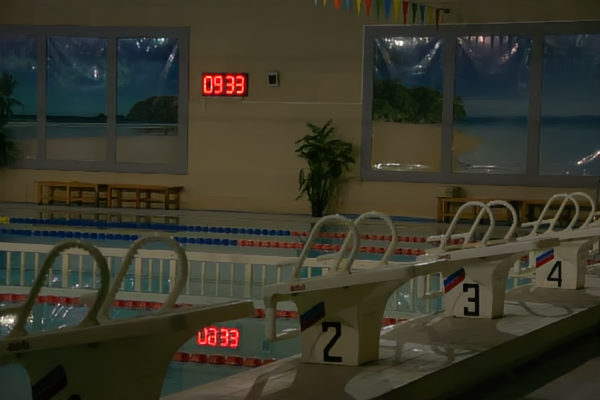} &

         \includegraphics[width=0.32\linewidth]{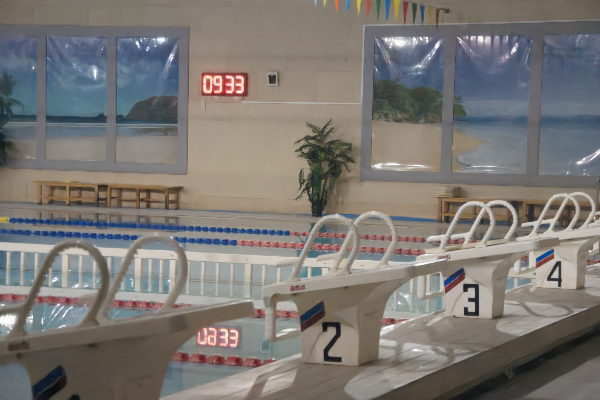} \\

         Low-light image without Blur & LEDNet~\cite{lednet} & \textbf{DarkIR} (Ours) \\
    \end{tabular}
    
    \captionof{figure}{Previous \textbf{Low-light Image Enhancement (LLIE)} and restoration methods are not robust to blur and illumination changes. Our \textit{multi-task model} is able to restore real low-light images under varying illumination, noise and blur conditions. Zoom-in to see details.}
    \label{fig:teaser}
    \vspace*{2.5mm}
\end{center}
}]

\begin{abstract}
Photography during night or in dark conditions typically suffers from noise, low light and blurring issues due to the dim environment and the common use of long exposure. Although Deblurring and Low-light Image Enhancement (LLIE) are related under these conditions, most approaches in image restoration solve these tasks separately. In this paper, we present an efficient and robust neural network for multi-task low-light image restoration. Instead of following the current tendency of Transformer-based models, we propose new attention mechanisms to enhance the receptive field of efficient CNNs. Our method reduces the computational costs in terms of parameters and MAC operations compared to previous methods. Our model, DarkIR, achieves new state-of-the-art results on the popular LOLBlur, LOLv2 and Real-LOLBlur datasets, being able to generalize on real-world night and dark images.
\end{abstract}

\section{Introduction}
\label{sec:intro}

During night or low-light conditions, we can define the image formation process as

\begin{equation}\label{eq:imagemodel}
  \mathbf{y}\!= \gamma\, (\mathbf{x}\otimes \mathbf{k})\, + \,\mathbf{n},
\end{equation}

\noindent where $\mathbf{y}$ is the observed dim image, $\mathbf{x}$ the unperturbed captured scene, $\mathbf{k}$ represents the lens (point-spread function) PSF blurring kernel, $\mathbf{n}$ is the additive sensor noise, and $\gamma$ is a function to control dynamic range and pixel saturation. We use $\otimes$ to represent the convolution operator.

In comparison with daytime conditions, at low-light, the noise (shot and read) is substantially higher. During night photography, cameras usually use long exposure (slow shutter speed) to allow more available light to illuminate the image. However, long exposure could lead to ghosting and blurring artifacts. These image degradations are more notable on smartphones, since these have a fixed aperture and limited optics. For these reasons, joint low-light enhancement and deblurring is paramount for mobile computational photography~\cite{lednet, delbracio2021mobile} --- see Figure~\ref{fig:teaser} samples.



Under any circumstances, the captured image will present certain levels of noise and blur~\cite{elad1997restoration}. Using a tripod to capture steady images, we can reduce notable the blur. Moreover, if we use proper illumination, we can notablely reduce the noise~\cite{hasinoff2014photon, abdelhamed2018high}.

However, nowadays most photographs are captured using (handheld) smartphones. Due to the limited sensor size and optics, these employ more complex Image Signal Processors (ISPs)~\cite{conde2022modelbased, delbracio2021mobile} in comparison to DSLM (Digital Single-Lens Mirrorless) cameras. Smartphone photography is still far from DSLM quality standards, yet recent research in low-level computer vision and computational photography is helping to close the gap. Low-light image enhancement (LLIE)~\cite{feng2024you, cai2023retinexformer, liu2021retinex, liu2024ntire, zhou2025glare, wang2023exposurediffusion}, image deblurring~\cite{deblurgan-v2, hu2014deblurring, gopro, realblur} and night photography enhancement~\cite{lednet, shutova2023ntire} are popular tasks.

For instance, avoiding the presence of blur due to hand-tremor (hand shaking) has been well-studied, even in low-light scenarios~\cite{liba2019handheld, zhao2016fast}. However, in most cases, these tasks are solved individually, thus, the state-of-the-art methods for image deblurring do not generalize on nighttime images, and the best methods for low-light enhancement cannot reduce the notable blur. This presents a clear limitation since multiple task-specific models need to be fine-tuned, stored, and applied in sequence, which limits their applications in real-world cases. 

To the best of our knowledge, very few works aim to solve these tasks (denoising, deblurring and LLIE) in a joint end-to-end manner~\cite{lednet, nbdn, zhao2022d2hnet, lv2024fourier}, being LEDNet~\cite{lednet} the most notable work. We focus on this research direction since exploiting the correlation between the degradations allows to achieve the best performance in terms of image reconstruction, usability and efficiency.

\vspace{-3mm}
\paragraph{Our contribution}
We propose a convolutional neural network (CNN) that operates both in the spatial and frequency domains. In the spatial domain, we focus on solving the noise $\mathbf{n}$ and the non-uniform blur $\mathbf{k}$, we achieve this by using large receptive field spatial attention. On the other hand, in the Fourier domain, we are able to enhance the low light conditions easily~\cite{li2023embedding, wang2023fourllie}, because of the global nature of the task. We can summarize our contributions as:

\begin{enumerate}

    \item{We design a lightweight neural network with frequency attention, and large receptive field attention, combining spatial and frequency information.}
    
    \item{Our model, DarkIR, achieves state-of-the-art results on the popular LOLBlur and Real-LOLBlur datasets~\cite{lednet}, improving \textbf{+1dB} in PSNR over LEDNet~\cite{lednet}, while having less computational cost.}

    \item DarkIR represents a new baseline for multi-task night/dark image enhancement.
    
\end{enumerate}

\section{Related Work}
\label{sec:related}

\paragraph{Image Deblurring} We can see decades of research on reconstructing sharp scenes. Reducing the blur in an image is divided into blind and non-blind methods. While the non-blind methods consider the blurring kernel $\mathbf{k}$ (or PSF) to process the image, the blind methods do not have any prior knowledge on the blur degradation.

In the recent years, multiple deep learning-based approaches have been proposed for blind and non-blind deblurring~\cite{gopro, deblurgan-v2}, surpassing in both scenarios the traditional methods. The non-blind approaches offer a great solution, considering that only blurry-sharp image pairs are required for training such models, and we do not require PSF estimation or any information about the sensor. Most of these approaches are sensor-agnostic \emph{i.e.,} they can enhance sRGB images captured from different cameras.

Nowadays a big part of these methods are based on convolutional neural networks (CNNs) \cite{generate_kernels, nah2017deep, deblurgan-v2, zamir2022restormer}. In DeblurGAN \cite{deblurgan-v2}, the authors use Generative Adversarial Networks (GANs) to solve this problem. More recently, the authors of \cite{kong2023efficient} implemented an efficient frequency domain based transformer for deblurring. We also find iterative methods and diffusion models \cite{lv2024fourier, Whang_2022_CVPR}.

\textbf{Low-Light Image Enhancement (LLIE)} The first methods used to consider image statistics or prior information \cite{abdullah2007dynamic, guo2016lime}, being most of them based on the well known Retinex Theory \cite{land1977retinex}. Following the deep learning tendency, nowadays LLIE methods are based in Convolutional Neural Networks (CNNs) such as RetinexNet~\cite{wei2018deep} (and the corresponding LOL dataset), ZeroDCE \cite{Zero-DCE} and SCI \cite{ma2022toward}.
Recent methods explore the power of transformers in this task, such in the case of RetinexFormer \cite{cai2023retinexformer}, or use the Fourier frequency information to enhance the amplitude of the image, like in FourLLIE~\cite{wang2023fourllie}. 

\textbf{Low-Light Blur Enhancement}. Even though image deblurring and low-light enhancement are tasks that capture great attention, solving both tasks at the same time is a challenging task, and very few works in the literature tackle it~\cite{zhao2022d2hnet, lednet, nbdn, lv2024fourier}. NBDN~\cite{nbdn} proposes a non-blind network to enhance night saturated images. When deconvolving the image to its sharp version, the presence of noise or saturated regions need to be kept in mind by the algorithm. With this work the authors proved that previous methods had issues solving this specific task. 

LEDNet \cite{lednet} aims to solve the problem of low-light enhancement considering that the images are also blurry. This is a realistic assumption since smartphones need long exposure times for the low-light environments. They develop an encoder-decoder network to solve this problem, and the popular LOLBlur and Real-LOLBlur datasets.


\begin{figure*}[t]
  \centering
  \includegraphics[width=\linewidth]{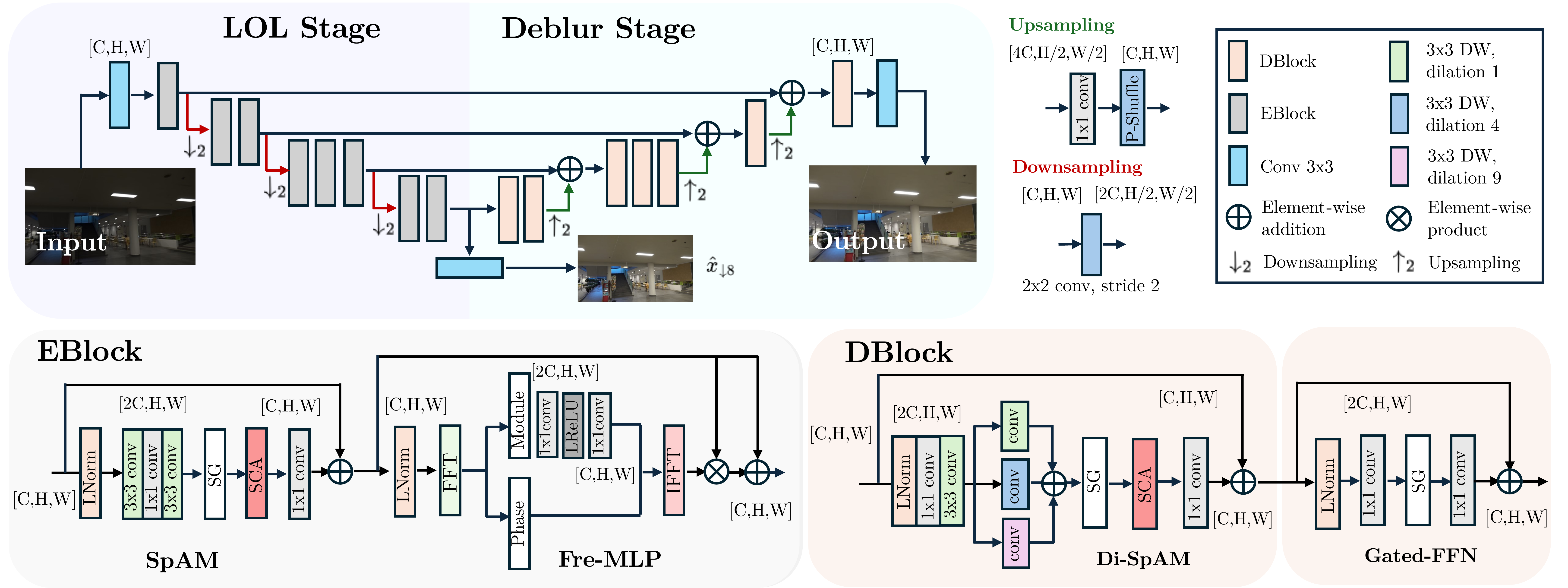}
  \caption{General diagram of \textbf{DarkIR}. The neural network follows an encoder-decoder architecture. We use different \textbf{blocks} for encoding and decoding that follow the Metaformer structure \cite{yu2022metaformer}. The encoder focuses on the low-light illumination issues using Fourier information. Thus, the encoder produces a low-resolution reconstructed image $\hat{x}_{\downarrow8}$ with corrected illumination. The decoder focuses on upscaling and reducing the blur using the prior illumination-enhanced encoded features. To achieve this, the decoder uses large receptive field spatial attention. This design allows our lightweight model to have less parameters and FLOPs than previous methods. 
  }
  \label{fig:general}
\end{figure*}

\section{Method}
\label{subsec:method}

We follow Metaformer \cite{yu2022metaformer} to design our neural network. This design simplifies transformer-based architectures into simple blocks with 2 components: global attention (\emph{e.g.,} token mixer), and a feed-forward network (FFN, MLP). The typical formulation for these blocks is:
\begin{align}
    z_1 = \text{Attention}(\text{LayerNorm} (z)) + z \\
    z_2 = \text{FFN}(\text{LayerNorm}(z_1)) + z_1
\end{align}

where $z$ are the input features and $z_2$ the output features of the block. Most popular image restoration models such as NAFNet~\cite{chen2022simple} use the same structure, and adapt the Attention module to different tasks.

We improve the metaformer structure by developing LLIE and deblurring specific blocks.

\paragraph{Low-light Enhancement} can be solved efficiently in the frequency domain. 

Many works~\cite{wang2023fourllie, li2023embedding} proved that the low-light conditions are highly correlated with the amplitude of the image in the Fourier domain. Thus, by enhancing just the amplitude of an image (without touching the phase), we can correct substantially the illumination of the image.
Moreover, this property stands at different resolutions~\cite{li2023embedding}. Therefore, we can estimate an illumination-enhanced image at low-resolution and upscale it.

\paragraph{Sharpening and Reducing Blur} usually require large receptive fields, this could be achieved by extracting deep features while downsampling the image -- NAFNet's approach~\cite{chen2022simple}. An alternative would be to use large kernels~\cite{luo2016understanding}, however, this could lead to more computational complexity and memory requirements.

\paragraph{DarkIR Model}
In Figure \ref{fig:general}, we illustrate our model. Unlike most previous methods, we use two different blocks for the encoder and decoder. The idea behind this asymmetry is to perform low-light enhancement at low-resolution in the encoder, and reduce the blur in the decoder, following a similar strategy as LEDNet \cite{lednet}. The decoder will use illumination-enhanced features from the encoder, and shall focus on upsampling and improve the sharpness of the already enhanced low-resolution reconstruction $\hat{x}_{\downarrow8}$.

For the encoder block, and to restore the low-light conditions, we work on the Fourier domain \cite{wang2023fourllie, li2023embedding}. The decoding block focuses on the spatial domain by incorporating dilated convolutions with large receptive field. By using task-specific blocks, we are able to use less blocks. This design allows to reduce notably the number of parameters and computational cost in terms of MACs and FLOPs.

\textit{How can we make sure the Encoder is performing low-light enhancement?} As we show in Figure \ref{fig:general}, the encoded features are linearly combined using a convolutional layer to produce an intermediate image representation $\hat{x}_{\downarrow8}$. We will use this to regularize our model using an additional loss function. By producing a good low-resolution representation, we can ensure that the amplitude in the Fourier domain has been properly enhanced.

\subsection{Low-light Enhancement Encoder}
\label{subsec:encoder}

We design the encoder blocks (\textbf{EBlock}) to enhance the low-light conditions of the image using Fourier information, and following the Metaformer~\cite{yu2022metaformer} (and NAFBlock~\cite{chen2022simple}) structure. The block has two components: the spatial attention module (SpAM) and a feed-forward network in the frequency domain (FreMLP). 

The spatial attention module resembles the NAFBlock~\cite{chen2022simple} with an inverted residual block followed by simplified channel attention (SCA). Instead of using activations, we use a simple gating mechanism, allowing our model to extract meaningful spatial information for enhancement in the frequency domain. 

As suggested by other works \cite{wang2023fourllie, lv2024fourier} the information related to the light conditions of the image depends mainly in the amplitude in the frequency domain. To enhance this, we apply the Fast Fourier Transform (FFT) and operate only over its amplitude. After this operation we transform again to the space domain with the Inverse Fast Fourier Transform (IFFT). The Fre-MLP serves as an additional attention mechanism. In this context, the MLP operating in the amplitude has better benefits than operating in the spatial domain (\emph{e.g., } channel MLP~\cite{yu2022metaformer}).

The encoder uses strided convolutions to downsample the features. After each level, the features have half of their original spatial resolution, which implies that more encoder blocks can be used in the deep levels without increasing notably the number of operations. 

Finally, the encoder will provide illumination-enhanced features to the decoder, and already a low-resolution estimation of clean image $\mathbf{x}$. This low-resolution image is $\hat{x}_{\downarrow8}$ estimated as a combination of the encoded deep features, it has a resolution $8\times$ smaller than the original one. Although it is a small estimation, the illumination (and amplitude) are consistent across scales~\cite{li2023embedding}.

\subsection{Deblurring Decoder}
\label{subsec:decoder}

The decoder block (\textbf{DBlock}) focuses on spatial transformations. The input of the decoder is deep representation of $\hat{x}_{\downarrow8}$, thus we can assume: (1) the decoder should focus on upsampling such an initial estimation, (2) the decoder should focus on reducing the blur and improving sharpness, since the illumination has been corrected by the encoder. In this block, we also maintain the metaformer structure~\cite{yu2022metaformer}: 

\begin{align}
    z_1 = \text{Di-SpAM}(\text{LayerNorm} (z)) + z \\
    z_2 = \text{GatedFFN}(\text{LayerNorm}(z_1)) + z_1
\end{align}

Inspired in Large Kernel Attention (LKA)~\cite{guo2023visual}, we create the Dilated-Spatial Attention Module (Di-SpAM). Unlike LKA~\cite{guo2023visual}, we use features at 3 different levels, using three dilated depth-wise convolutions with expand (dilation) factors 1,4,9. The attributes from the three branches are combined together, then we apply simplified channel attention to further enhance the features. Finally we use an MLP with simple gates instead of activations~\cite{chen2022simple}.

\subsection{Loss Function}
\label{subsec:loss}

Besides the new block designs, the loss function helped to maximize the potential of our approach. We use a combination of distortion losses and perceptual losses for optimizing our model $f$. First, to ensure high-fidelity (low distortion) we use $L_{pixel}$ defined as: $L_{pixel} = \lVert x - \hat{x} \rVert_1$, where $f(y) = \hat{x}$ and $x$ are respectively the enhanced and the ground-truth (clean) images. Thus, $L_{pixel}$ is the $\mathcal{L}_1$ loss.

To ensure high-fidelity we use $l1$ norm loss and for perceptual similarity we incorpore loss $L_{percep}$. For the last one we use LPIPS~\cite{zhang2018perceptuallpips} based on VGG19 \cite{simonyan2014very} to calculate the distance between features of our images: 

\begin{equation}
    L_{percep} = \text{LPIPS}\,(x, \hat{x}) .
\end{equation}

Using this loss we make sure that the network will produce a pleasant image, close to the clean reference. Following \cite{seif2018edge}, we also incorporate the gradient ($\nabla$) \emph{edge loss}: 

\begin{equation}
    L_{edge} = \lVert \nabla x - \nabla \hat{x} \rVert^2_2 
\end{equation}

that enforces consistency and accuracy in the reconstruction of the edges (high-frequencies).

Finally, similar to LEDNet~\cite{lednet}, we included $L_{lol}$ an architecture guiding loss to assert that the encoder focuses on the low-light enhancing. This loss works over the low-resolution output of the encoder $\hat{x}_{\downarrow8}$,

\begin{equation}
L_{lol}= \lVert x_{\downarrow8} - \hat{x}_{\downarrow8} \rVert_1
\end{equation}

comparing the intermediate result with the downsampled reference $x_{\downarrow8}$. Note that $\downarrow8$ indicates an $8\times$ resolution downsampling, obtained using bilinear interpolation.  

The complete loss function is then:

\begin{equation}
    \mathcal{L} = \lambda_p\cdot L_{l1} + \lambda_{pe} \cdot L_{percep} + \lambda_{ed} \cdot L_{edge} + L_{lol} 
\end{equation}

The constants $\lambda_p$, $\lambda_{pe}$, and $\lambda_{ed}$ are loss weights empirically set to 1, $1e^{-2}$, and 50 respectively. 

\begin{table*}[t]
  \caption{\textbf{Quantitative evaluation on the LOLBlur dataset.} DarkIR achieves new state-of-the-art results in distortion and perceptual metrics. Moreover, we have $55\%$ less parameters than LEDNet\cite{lednet} and $88\%$ less than Restormer \cite{zamir2022restormer}, which is key in memory-constrained devices. This table --specially numbers for previous methods retrained in LOLBlur-- recovers results on previous analysis~\cite{lednet} and adds new retrained ones. Best and second best results are bolded and underlined, respectively.
  }
  \label{tab:lolblur-results}
  \centering
  \resizebox{\textwidth}{!}{%
  \begin{tabular}{l ccccccccccc}
    \toprule
    &KinD++\cite{kind++} & DRBN\cite{drbn} & DeblurGAN-v2\cite{deblurgan-v2} & MIMO\cite{mimo} & NAFNet\cite{chu2022nafssr} & LEDNet\cite{lednet} & RetinexFormer\cite{cai2023retinexformer} & Restormer\cite{zamir2022restormer} & DarkIR-m (Ours) & DarkIR-l (Ours) \\
    \midrule
    PSNR (dB) $\uparrow$ & 21.26 & 21.78 & 22.30  & 22.41 & 25.36 & 25.74 & 26.02 & 26.72 & \underline{27.00} & \textbf{27.30} \\
    
    SSIM $\uparrow$ & 0.753 & 0.768 & 0.745 & 0.835  & 0.882 & 0.850 & 0.887 & \textbf{0.902} & 0.883 & \underline{0.898} \\
    
    LPIPS $\downarrow$ & 0.359 & 0.325 & 0.356 & 0.262 & 0.158 & 0.224 & 0.181 & \textbf{0.133} & 0.162 & \underline{0.137} \\
    
    Params (M) $\downarrow$ & \underline{1.2} & \textbf{0.6} & 60.9 & 6.8 & 12.05 & 7.4 & 1.61 & 26.13 & 3.31 & 12.96 \\
    MACs (G) $\downarrow$ & 34.99 & 48.61 & - & 67.25 & \underline{12.3} & 38.65 & 15.57 & 144.25 & \textbf{7.25} & 27.19 \\
  \bottomrule
  \end{tabular}}
\end{table*}

\section{Experimental Results}
\label{sec:experimental}

\subsection{Datasets}
\label{sec:datasets}

To train our model and evaluate its ability to reconstruct low-light blurred images, we use the LOLBlur dataset~\cite{lednet}. Although there are other datasets for this task such as NBDN~\cite{nbdn}, LOLBlur offers a large-scale synthetic dataset produced using a sophisticated pipeline.

\vspace{-2mm}
\paragraph{LOLBlur} has 10200 training pairs, and 1800 testing pairs. Note that this is a \emph{synthetic dataset}, although generated in a realistic manner~\cite{lednet}: the data is generated by averaging frames to synthetize blur and darkening the normal-light images with EC-Zero-DCE (a variant of Zero-DCE \cite{Zero-DCE}). We use the dataset variant that includes real sensor noise, which makes our model more effective in challenging conditions and real scenarios.

\vspace{-2mm}
\paragraph{Real-LOLBlur} is a \emph{real-world test} dataset, that contains 482 real-world night blurry images selected from RealBlur~\cite{realblur} to verify the generalization of the proposed method. Note that these images do not have a ground-truth since these were captured in the wild. 

\vspace{-2mm}
\paragraph{LOLv2 (real)} is a real-world dataset that includes 689 low/high paired images for training and 100 low/high paired images for testing. Note that LOLv2-Real \cite{yang2021sparse} is the extended version of LOL\cite{Chen2018Retinex}, thus, we use the v2 version directly. We also use \emph{LOLv2-Synthetic} that includes 900 pairs of low/high images for training and 100 validation ones.

\vspace{-2mm}
\paragraph{LSRW} includes images from a DSLM Nikon camera and a Huawei smartphone. The LSRW-Nikon dataset is composed of 3150 training image pairs and 20 testing image pairs. The LSRW-Huawei dataset contains 2450 pairs of images and 30 pairs for training and validation, respectively.

\subsection{Results}
\label{seq:results}

\begin{figure*}[t]
  \centering
  \includegraphics[width=\textwidth]{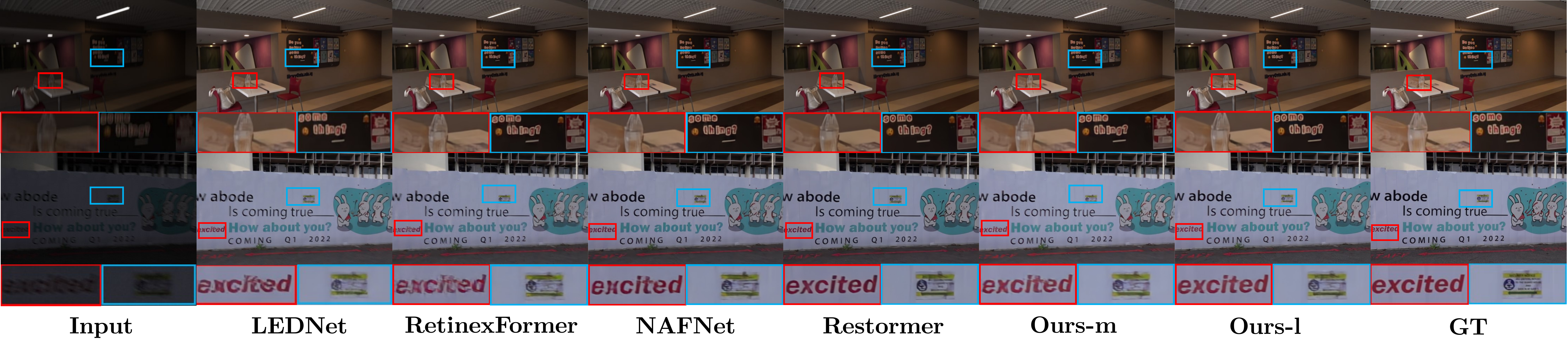}
  \vspace{-5mm}
  \caption{Qualitative comparisons on the \textbf{LOLBlur} dataset (synthetic samples from the testset).
  }
  \label{fig:LOLBlur}
\end{figure*}

We quantitatively and qualitatively evaluate the proposed DarkIR on the LOLBlur Dataset~\cite{lednet}. Implementations details can be found in the supplementary. We retrained some general purpose state-of-the-art methods in LOLBlur and compare them with LEDNet~\cite{lednet} and our proposed network. We follow the baseline methods analyzed in previous works: (1,2) zero-shot methods trained for real-world cases, and (3) fine-tuned methods for this particular task.

\vspace{2mm}
\noindent
\textbf{1. LLIE $\rightarrow$ Deblurring.}
We consider Zero-DCE~\cite{Zero-DCE}, RUAS~\cite{liu2021retinex} and RetinexFormer~\cite{cai2023retinexformer} as LLIE models, followed by a deblurring network MIMO-UNet~\cite{mimo} or NAFNet~\cite{chen2022simple}.

\vspace{2mm}
\noindent
\textbf{2. Deblurring $\rightarrow$ LLIE.}
For deblurring, we include popular baselines such as DeblurGAN-v2~\cite{deblurgan-v2} trained on the RealBlur~\cite{realblur} dataset, and MIMO-UNet~\cite{mimo} or NAFNet~\cite{chen2022simple} trained on GoPro~\cite{gopro} dataset.
%
We employ Zero-DCE~\cite{Zero-DCE} and RetinexFormer~\cite{cai2023retinexformer} for light enhancement. 

\vspace{2mm}
\noindent
\textbf{3. End-to-end training on LOLBlur dataset.}
We consider the following LLIE models re-trained on the LOLBlur dataset: KinD++~\cite{kind++}, DRBN\cite{drbn} and RetinexFormer\cite{cai2023retinexformer}. In addition, we consider four deblurring networks: DeblurGAN-v2~\cite{deblurgan-v2}, NAFNet\cite{chu2022nafssr}, MIMO-UNet~\cite{mimo} and Restormer\cite{zamir2022restormer}.

\vspace{2mm}
\noindent\textbf{4. Multi-Task Results.} We also trained our model for \emph{practical low-light restoration} by combining LOLBlur, LOLv2 and LSRW datasets (following all-in-one restoration methods~\cite{conde2024high, potlapalli2023promptir, Li_2022_CVPR, zhang2023ingredient}). Results can be seen in Tables \ref{tab:v2real} and \ref{tab:lsrw}, where the all-in-one training showcases promising results. As a multi-task method, \textbf{DarkIR-mt} outperforms previous LLIE methods, while being also robust to blur (previous methods are only robust to illumination and noise). However, the performance on LOLBlur is 26.62 dB, suffering a slight -0.4dB loss. Figures \ref{fig:v2real} and \ref{fig:lsrw} showcase these results in LOLv2-Real and LSRW  datasets, respectively.

\begin{figure*}[t]
	\begin{center}
		\includegraphics[width=\linewidth]{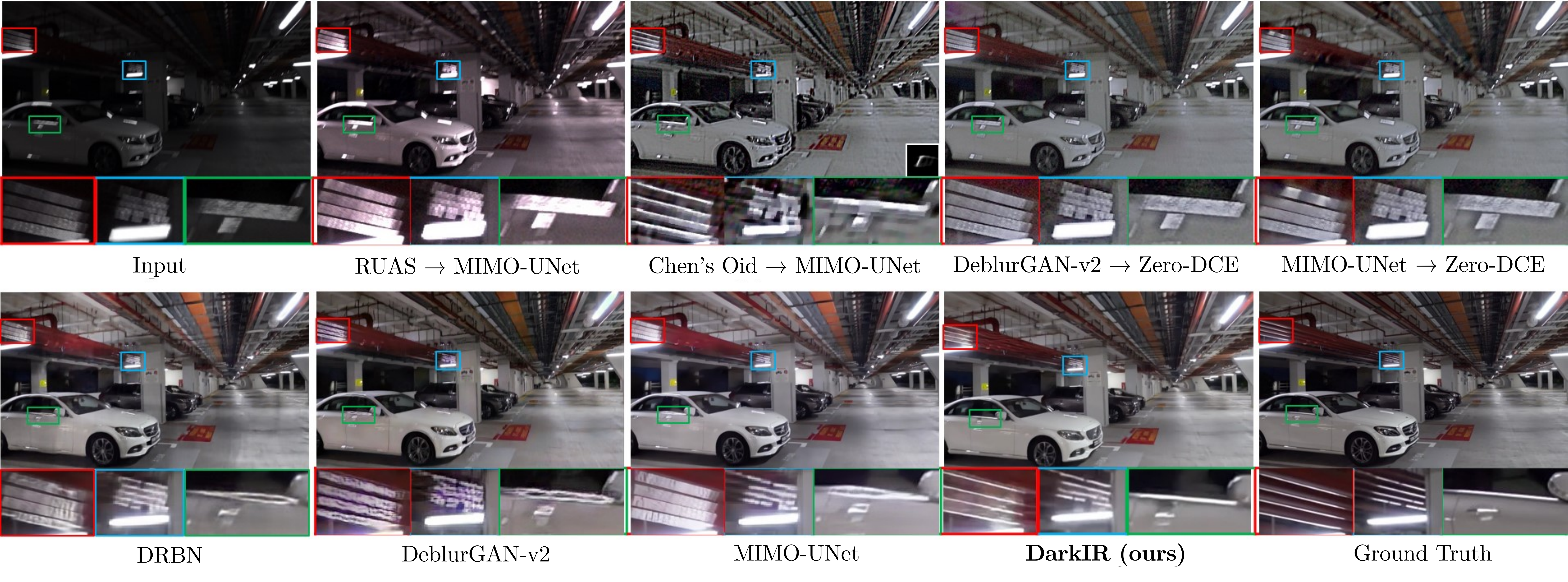}
        \vspace{-5mm}
		\caption{{Additional visual comparisons on the \textbf{LOLBlur}~\cite{lednet} dataset with 2-step pipelines. DarkIR generates much sharper images with visually pleasing results. (Zoom in for best view}).}
		\label{fig:lolblur_others}
	\end{center}
\end{figure*}
%

\vspace{-2mm}
\paragraph{Evaluation Metrics.}
We employ traditional quality (distortion) metrics PSNR and SSIM for evaluation on the synthetic LOLBlur dataset. To evaluate the perceptual quality of the restored images, we use the perceptual metric LPIPS~\cite{zhang2018perceptuallpips} between the reference and reconstructed image.

\vspace{-2mm}
\paragraph{Quantitative and Qualitative Results}

In Table~\ref{tab:lolblur-results} we compare with fine-tuned methods for this task. We improve previous state-of-the-art LEDNet~\cite{lednet} by $+1$db in terms of PSNR, and we reduce LPIPS by half. Table \ref{tab:lolblur-results} also showcases that DarkIR performs better than most of the other general purpose methods, while reducing params and computing cost. Figure~\ref{fig:LOLBlur} supports qualitatively these results.  

 In Table~\ref{tab:lolblur_others} we compare with 2-step (zero-shot) pipelines, with a clear dominance of DarkIR. Figure~\ref{fig:lolblur_others} compares DarkIR with 2-step pipelines, were it shows more details and sharpness. We provide more results in the appendix. 

\begin{table*}[t]
	\centering
	\caption{{Additional quantitative evaluation on LOLBlur dataset with enhancement pipelines (Deblurring + LLIE).
		}}
	\resizebox{0.9\linewidth}{!} {
		\begin{tabular}{lccc | cccc | c}
		\toprule
  
		\multirow{3}{*}{} & \multicolumn{3}{c|}{\textbf{1. LLIE $\rightarrow$ Deblurring}} & \multicolumn{4}{c|}{\textbf{2. Deblurring $\rightarrow$ LLIE}} & 
        \\

        \midrule
  
		
        & Zero-DCE~\cite{Zero-DCE} & RUAS~\cite{liu2021retinex}  & RetinexFormer \cite{cai2023retinexformer} & Chen~\cite{chen2021blind} & DeblurGAN-v2~\cite{deblurgan-v2} & MIMO~\cite{mimo} & NAFNet~\cite{chen2022simple} & \textbf{DarkIR-m (Ours)} 
        \\

		& $\rightarrow$ MIMO~\cite{mimo} & $\rightarrow$ MIMO~\cite{mimo} & $\rightarrow$ NAFNet~\cite{chen2022simple} & $\rightarrow$ Zero-DCE~\cite{Zero-DCE} & $\rightarrow$  Zero-DCE~\cite{Zero-DCE}  & $\rightarrow$  Zero-DCE~\cite{Zero-DCE} & $\rightarrow$ RetinexFormer~\cite{cai2023retinexformer} & End-to-End 
        \\
		\midrule
  
        PSNR (dB) $\uparrow$ & 17.68 & 17.81 & 17.16   & 17.02 & 18.33 & 17.52 & 14.66 & \textbf{27.00} 
        \\
        
		SSIM $\uparrow$ & 0.542  & 0.569 & 0.673   & 0.502 & 0.589 & 0.57 & 0.500 & \textbf{0.883} 
        \\
	
        LPIPS $\downarrow$  & 0.510 & 0.523& 0.392   & 0.516 & 0.476 & 0.498 & 0.465 & \textbf{0.162} 
        \\

        \bottomrule
		\end{tabular}
	}
	\label{tab:lolblur_others}
\end{table*}

\begin{table}[t]
\caption{Results on LOLv2-Real~\cite{yang2021sparse} and LOLv2-Synthetic~\cite{yang2021sparse}. Our \emph{multi-task model (DarkIR-mt)} obtains new SOTA results by leveraging all-in-one training, and including low-light deblurring. The base model trained only for LOL (DarkIR-lol) also achieves SOTA results, which proves the efficacy of the architecture. Table based on~\cite{cai2023retinexformer, wang2023fourllie}. MACs were calculated on $256\times256\times3$.} 

\label{tab:quantitative}
\begin{adjustbox}{width=1\columnwidth,center}
	\centering
	\setlength\tabcolsep{4pt}
	\resizebox{\linewidth}{!}{\hspace{-0.5mm}
		\begin{tabular}{r cc cc cc cc }
			\toprule[0.15em]
			\multirow{2}{*}{Methods}      & \multicolumn{2}{c}{Complexity} & \multicolumn{2}{c}{LOLv2-Real}  &\multicolumn{2}{c}{LOLv2-Syn}   \\  & MACs (G)$\downarrow$ & Params (M)$\downarrow$ & PSNR $\uparrow$ & SSIM $\uparrow$ & PSNR $\uparrow$ & SSIM $\uparrow$ \\
            \midrule[0.15em]

			UFormer~\cite{uformer}   &12.00 &5.29        & 18.82        & 0.771        & 19.66     &0.871   \\
			
			RetinexNet~\cite{Chen2018Retinex}  & 587.47    & 0.84   & 15.47        & 0.567  &17.13  &0.798  \\
			
			EnGAN~\cite{enlightengan}  &61.01  &114.35    &18.23  &0.617   &16.57   &0.734 \\
			
			RUAS~\cite{liu2021retinex}     &\bf{0.83} &\bf{0.003}  &18.37  &0.723   &16.55    &0.652    \\
			
			FIDE~\cite{xu2020learning}     &28.51 &8.62  &16.85  &0.678   &15.20    &0.612  \\
			
			DRBN~\cite{drbn} &48.61  &5.27    &20.29   & 0.831    & 23.22     & 0.927   \\
			
			KinD~\cite{kind++} &34.99    &8.02 &14.74 &0.641  &13.29 &0.578  \\
			
			Restormer~\cite{zamir2022restormer}   &144.25 &26.13       &19.94         &0.827         &21.41      &0.830   \\

			MIRNet~\cite{mirnet_v2}   &785 &31.76  &20.02   &0.820  &21.94  &0.876  \\
   
			SNR-Net~\cite{snr_net}   &26.35   &4.01   &21.48  &\underline{0.849}  &24.14 & 0.928   \\

            FourLLIE~\cite{wang2023fourllie}      &\underline{5.8}     & \underline{0.120}    &21.60   & 0.847    &24.17      &0.917      \\
   
			Retinexformer~\cite{cai2023retinexformer}      &15.57  &1.61        &\underline{22.80}      &0.840    &\bf{25.67} &\underline{0.930} \\
    
            \midrule[0.15em] 
            
            \textbf{DarkIR-mt (Ours)}     & 7.25     & 3.31   &\bf{23.87}   & \bf{0.880} &\underline{25.54}      & \textbf{0.934}    \\
            \bottomrule[0.15em]
	\end{tabular}}
    \end{adjustbox}
	\vspace{2mm}
    \label{tab:v2real}
	\vspace{-3mm}
\end{table}

\begin{table}[t]
    \centering
    \caption{Metrics on the LSRW dataset (50 test images from Huawei and Nikon)~\cite{hai2023r2rnet}. All the values are adopted from \cite{Yang_2023_ICCV,ma2022toward}.
    }

    \resizebox{1\linewidth}{!}{
    \begin{tabular}{c ccccc}
    \toprule
    
    & RetinexNet & FIDE & DRBN & KinD & STAR \\
    & \cite{Chen2018Retinex} & \cite{xu2020learning} & \cite{drbn} & \cite{kind++} & \cite{xu2020star} \\
    \midrule
    PSNR $\uparrow$ & 15.906 & \underline{17.669} & 16.149 & 16.472 & 14.608 \\
    SSIM $\uparrow$ & 0.3725 & \underline{0.5485} & 0.5422 & 0.4929 & 0.5039 \\
    \midrule
    & EnGAN & ZDCE & RUAS & SCI & \bf{DarkIR-mt} \\
    & \cite{enlightengan} & \cite{Zero-DCE} & \cite{liu2021retinex} & \cite{ma2022toward} & (Ours) \\
    \midrule
    PSNR $\uparrow$ & 16.311 & 15.834 & 14.437 & 15.017 & \bf{18.93}  \\
    SSIM $\uparrow$ & 0.4697 & 0.4664 & 0.4276 & 0.4846 & \bf{0.583}  \\

    \bottomrule
    \end{tabular}
    }
    \label{tab:lsrw}

\end{table}

\begin{table}[t]
    \centering
    \caption{{Perceptual quality metrics on Real-LOLBlur~\cite{lednet}.}}
    \setlength\tabcolsep{2pt}
    \resizebox{1\linewidth}{!}{
        \begin{tabular}{l c c c c c c c c}
            \toprule
            & RUAS & MIMO & RetinexFormer & NAFNet &  LEDNet & Restormer  & \textbf{DarkIR-m} & \textbf{DarkIR-l }    \\
            
            & $\rightarrow$ MIMO & $\rightarrow$ Zero-DCE &           &            &          & & &            \\
            
            \midrule
            
            MUSIQ$\uparrow$  & 34.39              & 28.36                  & 45.30        & \textbf{50.22} & 39.11 & 46.6 & 48.36 & \underline{48.79} \\ 
            
            \midrule
            
            NRQM$\uparrow$   & 3.322              & 3.697                  & 5.281        & 4.940 & \bf{5.643} & 4.627 & \underline{4.983} & 4.917 \\ 
            
            \midrule
            
            NIQE$\downarrow$ & 6.812              & 6.892                  & 4.576         & \underline{5.123} & 4.764 & \bf{5.268} & 4.998 & 5.051 \\ \bottomrule
        \end{tabular}
    }
    \label{tab:real_test}
\end{table}


\subsection{Evaluation on Real Data}

We use the \textbf{Real-LOLBlur}~\cite{lednet} to evaluate the robustness of our method on real-world cases. Since there is no ground-truth for the test images, we use well-known blind quality assessment metrics. We present results in other \emph{real-world (unpaired) LLIE datasets} in the appendix.

\noindent\textbf{Evaluation Metrics.}
We employ the recent image quality assessment methods: MUSIQ~\cite{ke2021musiq}, NRQM~\cite{ma2017learning} and NIQE~\cite{mittal2012making} as our perceptual metrics. Following previous works, we choose the MUSIQ model trained on KonIQ-10k dataset, which focuses more on color contrast and sharpness assessment -- quite suitable for our task. We use the pyiqa~\footnote{\url{https://pypi.org/project/pyiqa/}} implementation of these metrics.

\noindent\textbf{Quantitative Evaluations.}
As shown in Table~\ref{tab:real_test}, the proposed DarkIR achieves competitive perceptual quality scores in terms of the three perceptual metrics, indicating that our method performs in tune with human perception. 

\vspace{2mm}

\begin{figure*}[!ht]
  \centering
  \includegraphics[width=\textwidth]{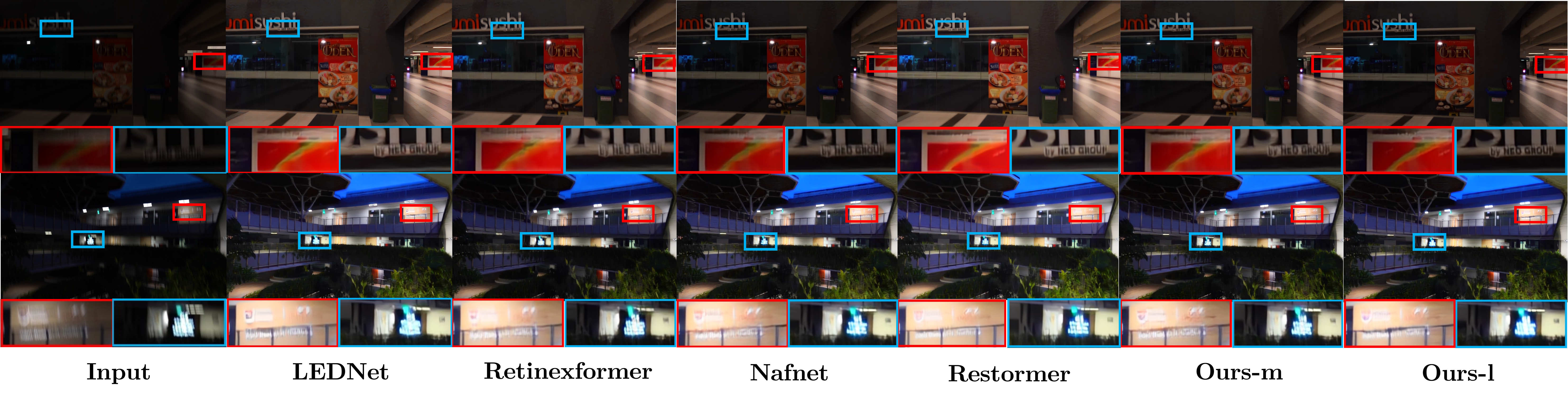}
  \caption{Qualitative comparison on real night scenes from the \textbf{RealBlurLOL} dataset. 
  }
  \label{fig:RealBlur1}
\end{figure*}

\noindent\textbf{Qualitative Evaluations.}
Fig.~\ref{fig:RealBlur1} presents visual comparisons on real-world night blurry image from Real-LOLBlur~\cite{lednet}.
These samples showcase the robustness of our approach on real cases with handheld motion blur, sensor noise, saturated pixels and low illumination. We provide more results in the supplementary material.

\begin{figure*}[t]
    \centering
    \setlength{\tabcolsep}{1pt} 
    \resizebox{0.98\linewidth}{!}{
    \begin{tabular}{c c c c c c}

    \includegraphics[width=0.165\linewidth]{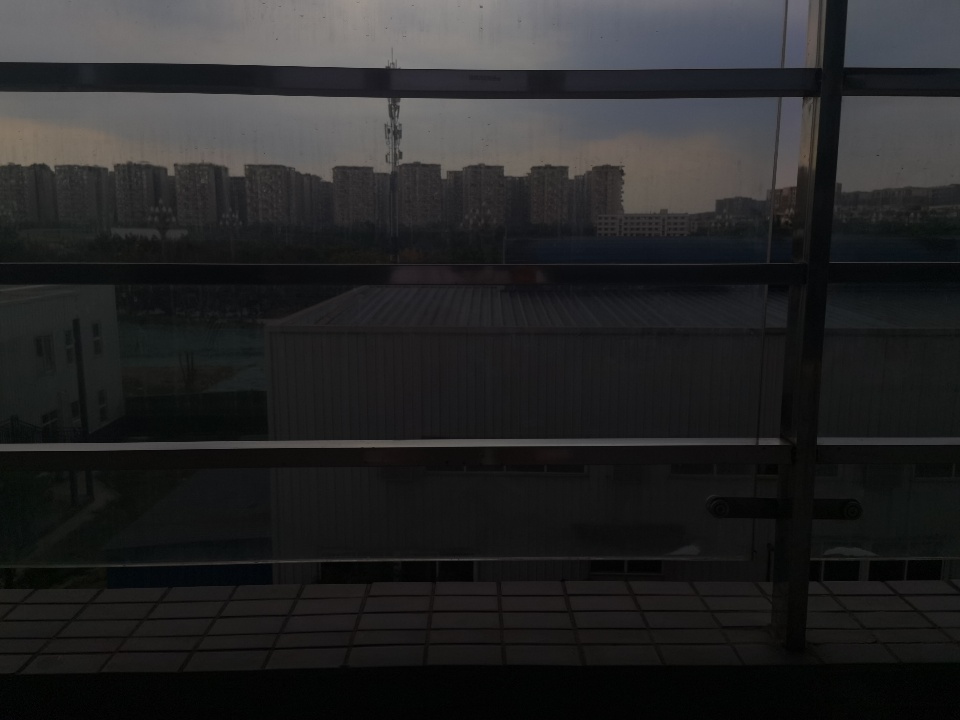} &

    \includegraphics[width=0.165\linewidth]{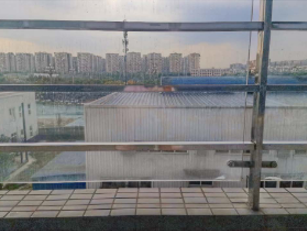} &

    \includegraphics[width=0.165\linewidth]{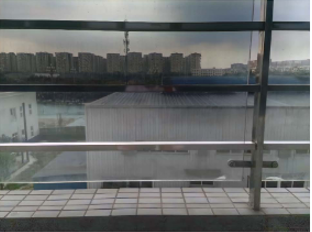} &

    \includegraphics[width=0.165\linewidth]{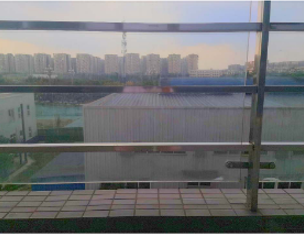} &

    \includegraphics[width=0.165\linewidth]{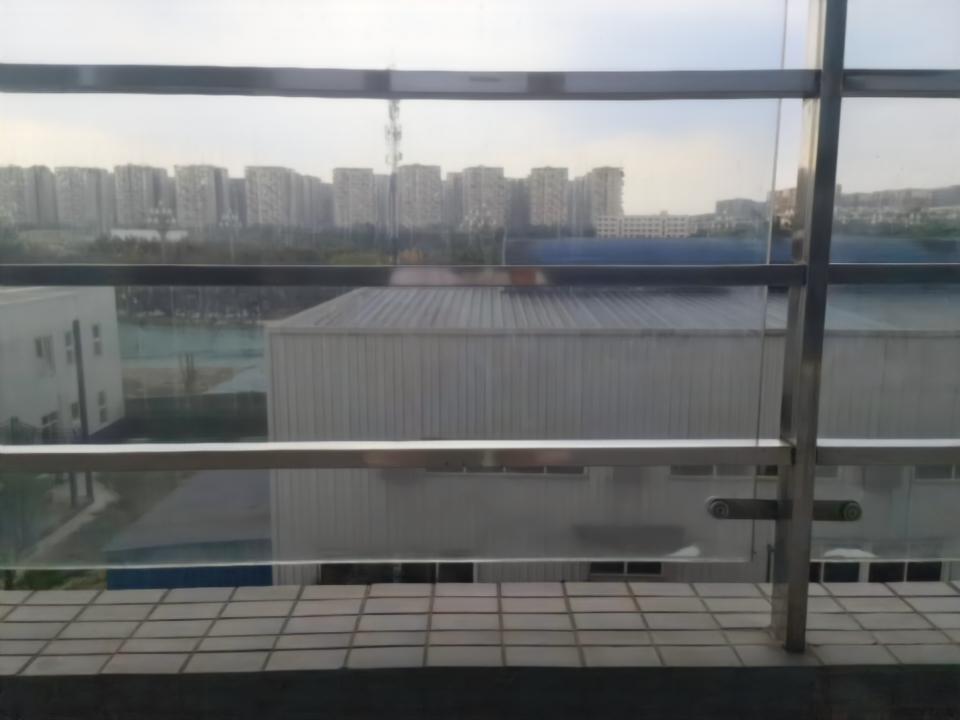} &

    \includegraphics[width=0.165\linewidth]{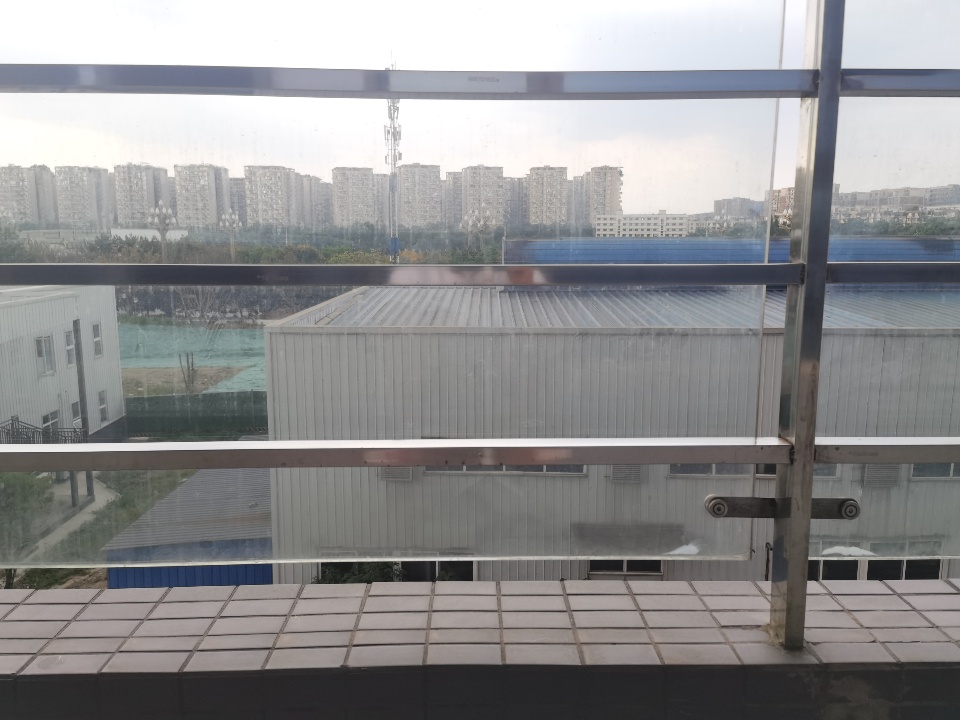} \\

    Input & KinD & DRBN & EnGAN & \bf{DarkIR-m} & Ground Truth \\

    \end{tabular}
    }
    \caption{Qualitative results on the real-world dataset \textbf{LSRW-Huawei}~\cite{hai2023r2rnet}. We provide more samples in the supplementary.}
    \label{fig:lsrw}
\end{figure*}

\begin{figure*}[t]
    \centering
    \setlength{\tabcolsep}{1pt} 
    \begin{tabular}{c c c c }

    \includegraphics[width=0.24\linewidth]{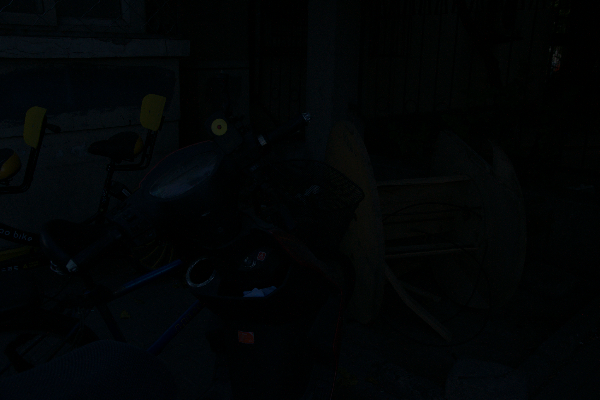} &

    \includegraphics[width=0.24\linewidth]{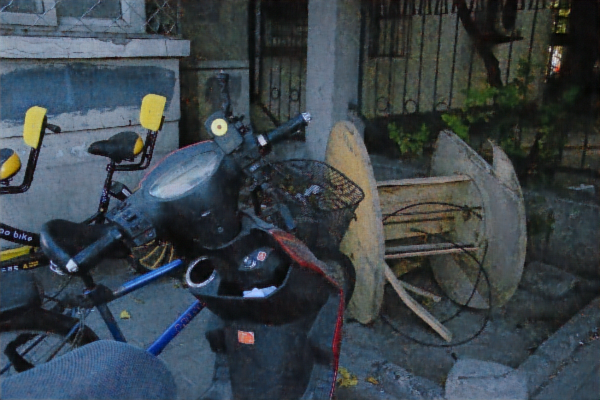} &

    \includegraphics[width=0.24\linewidth]{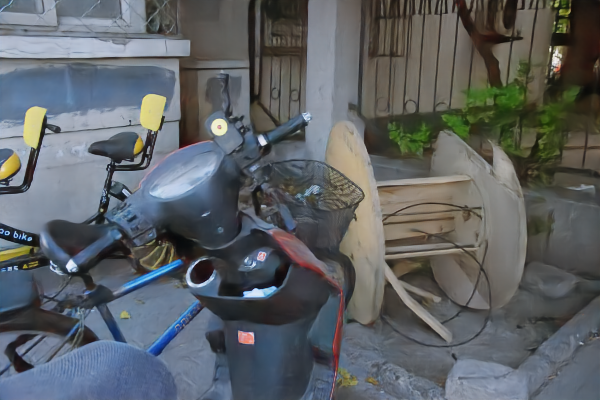} &

    \includegraphics[width=0.24\linewidth]{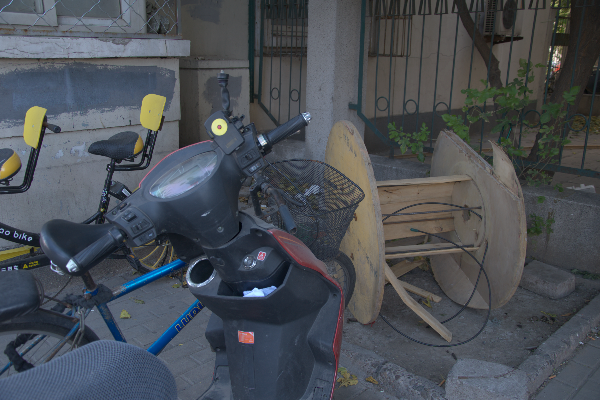} \\

    \includegraphics[width=0.24\linewidth]{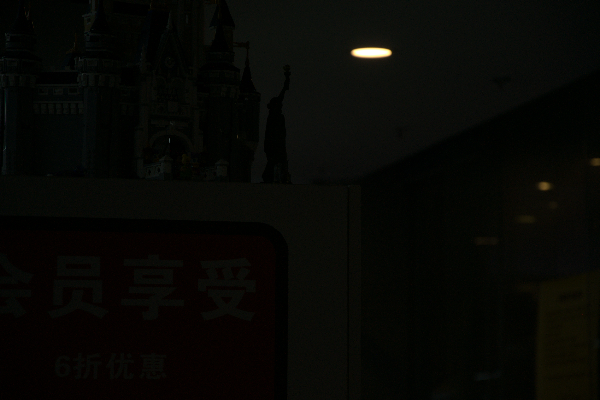} &

    \includegraphics[width=0.24\linewidth]{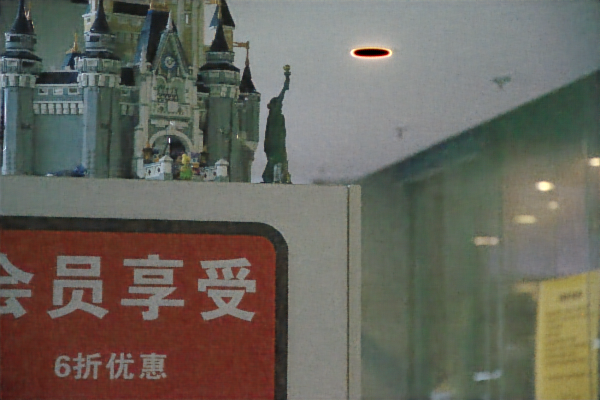} &

    \includegraphics[width=0.24\linewidth]{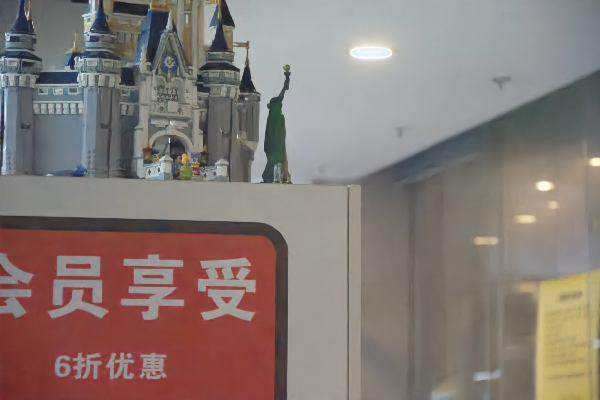} &

    \includegraphics[width=0.24\linewidth]{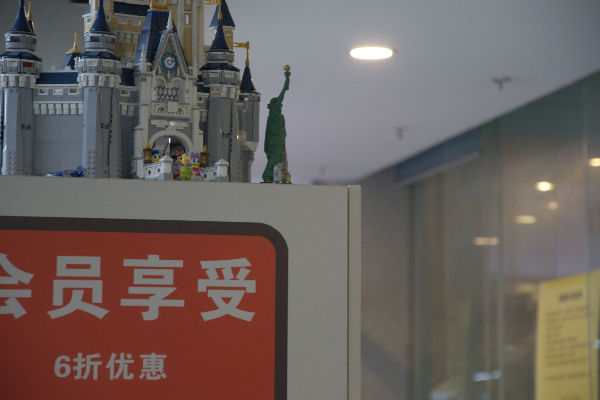} \\

    Input & RetinexFormer~\cite{cai2023retinexformer} & \textbf{DarkIR-m} & Ground Truth \\

    \end{tabular}
    \caption{Qualitative results compared with state of the art method RetinexFormer \cite{cai2023retinexformer} on \textbf{LOLv2-Real}.}
    \label{fig:v2real}
\end{figure*}

\subsection{Ablation Study}
\label{sec:ablation}

In addition to our results, we include three ablation studies on the model’s design and training.

In Table~\ref{tab:abla_block} we compare results obtained using different block configurations, like NAFBlock, EBlock or DBlock. The first row  specifies the change applied to the network, while everything else remains exactly as in the proposed model. We can clearly see how our model achieves the best results. Unlike the final proposed model, all the models in this study were trained using crops of 256px instead of 384px, which explains the lower results in general. We also studied the decoder block spatial attention in Table~\ref{tab:abla_lka}, were we compare the well-known Large Kernel Attention (LKA) mechanism, with our Dilated-spatial Attention Module (Di-SpAM). Our approach performs better and requires less parameters and operations.

In Table~\ref{tab:abla_channels} we also explore the scalability of our model by changing its channel embedding. As expected, by increasing the embedding size, the performance raises. In ascending order of parameters, we have channel embeddings of 16, 32, and 64. We find the best efficiency/performance balance in the 32 channels embedding (\textbf{DarkIR-m}).

In the supplementary material, we present additional studies from the architecture’s development.

\begin{table}[t]
    \centering
    \caption{Network blocks ablation study. Combination of EBlocks and DBlocks achieves the best performance.}
    \setlength\tabcolsep{2pt}
    \resizebox{1\linewidth}{!}{
        \begin{tabular}{l c c c c c}
            \toprule
                             & Params$\downarrow$ (M)   & MACs$\downarrow$ (G)     & PSNR$\uparrow$ & SSIM$\uparrow$ & LPIPS$\downarrow$ \\\midrule
            EBlock is NAFBlock  & 3.12             & 7.69                  & 26.51     & 0.8764    & 0.169 \\ 
            EBlock has also Phase Transform  & 4.08             & 8.38                  & 26.30     & 0.871    & 0.179 \\ 
            All DBlock   & 3.2              & 8.19                  & 26.68     & 0.876    & 0.170   \\ 
            All EBlock  & 3.44             & 6.46                  & 26.17     & 0.863    & 0.187 \\ 
            All NAFBlock~\cite{chen2022simple} & 3.04             & 7.29                  & 26.24     & 0.868    & 0.178 \\ 
            DBlock is NAFBlock  & 3.24             & 6.84                  & 26.23     & 0.864    & 0.185 \\ 
            DBlock w/o Extra Depthwise  & 3.27             & 6.99                 & 26.63     & 0.875    & 0.174 \\ \midrule
            \textbf{DarkIR-m}  & 3.31             & 7.25                  & 26.90     & 0.874    & 0.175 \\ 
            \bottomrule

        \end{tabular}
    }
    \label{tab:abla_block}
\end{table}

\begin{table}[t]
    \centering
    \caption{{Spatial Attention ablation study. We compare LKA (large kernel attention)~\cite{guo2023visual} with our proposed Di-SpAM for the decoder blocks. MACs were calculated considering an input of 256px.}}
    \setlength\tabcolsep{2pt}
    \resizebox{1\linewidth}{!}{
        \begin{tabular}{l c c c c c}
            \toprule
                             & Params$\downarrow$ (M)   & MACs$\downarrow$ (G)     & PSNR$\uparrow$ & SSIM$\uparrow$ & LPIPS$\downarrow$ \\\midrule
            LKA~\cite{guo2023visual}  & 4.06             & 9.14                  & 26.45     & 0.876    & 0.172 \\ \midrule
            Di-SpAM (DarkIR-m)   & 3.31              & 7.25                  & 27.00     & 0.883    & 0.162   \\ \bottomrule

        \end{tabular}
    }
    \label{tab:abla_lka}
\end{table}

\begin{table}[t]
    \centering
    \caption{{Ablation study scaling the channel depth dimensions. We can appreciate how our model scales properly, which allows adaptation depending on memory or runtime requirements.}}
    \setlength\tabcolsep{2pt}
    \resizebox{1\linewidth}{!}{
        \begin{tabular}{r c c c c c}
            \toprule
                             & Params$\downarrow$ (M)   & MACs$\downarrow$ (G)     & PSNR$\uparrow$ & SSIM$\uparrow$ & LPIPS$\downarrow$ \\\midrule
            DarkIR-s (16)  & 0.872            & 2.04           & 26.15     & 0.857    & 0.206 \\ \midrule
            DarkIR-m (32)  & 3.31             & 7.25                  & 27.00     & 0.883    & 0.162 \\ \midrule
            DarkIR-l (64)   & 12.96              & 27.19                & 27.30     & 0.898    & 0.137   \\ \bottomrule

        \end{tabular}
    }
    \label{tab:abla_channels}
\end{table}

\subsection{Efficiency Discussion}
\label{sec:efficiency}

By using the EBlock and DBlock we are able to reduce greatly the number of parameters of the network, getting an outstanding \textbf{$55\%$ less parameters} than LEDNet~\cite{lednet} (previous state-of-the-art) and $88\%$ less than Restormer \cite{zamir2022restormer} (second best method). This reduction is also accompanied by a reduction in the number of operations needed to enhance the input image. Considering an image of 256px -as previous works~\cite{chen2022simple}-, LEDNet uses 33.74 GMACs (Multiply-Accumulate Operations) and Restormer 141.24 GMACs, while DarkIR only uses \textbf{7.25 GMACs} -- note that 1 MAC is roughly 2 FLOPs. This means a reduction of $4\times$ the number of operations with previous state-of-the-art method and almost $20\times$ to the second best one. 

Therefore, DarkIR, while being state-of-the-art in low-light deblurring, is also lighter in all aspects, representing an advancement towards deploying this kind of models on devices with low computational power.

\section{Limitations}
Although we are able to reduce the computational requirements of the target device for running our model, we have done this by using depth-wise convolutions, which are not necessarily optimal in certain GPUs architectures, as they lack of arithmetic intensity \cite{gholami2018squeezenext}. Due to this, the model's inference times are not reduce drastically and proportionally with the reduction in operations. As future work, we will propose new methods that can combine the low computational requirements with notably faster inference times.

\section{Conclusion}

We propose a model for multi-task low-light enhancement and restoration. Our model, DarkIR, is an efficient and robust neural network that performs denoising, deblurring and low-light enhancement on dark and night scenes. DarkIR, achieves new state-of-the-art results on the popular LOLBlur, LOLv2 and Real-LOLBlur datasets, being able to generalize on real-world night blurry images while being more efficient than previous methods. 


\input{suppl}


\clearpage

{
    \small
    \bibliographystyle{ieeenat_fullname}
    \bibliography{refs}
}

\end{document}

%% file: suppl.tex
\clearpage
\setcounter{page}{1}
\maketitlesupplementary

\setcounter{section}{0}
\setcounter{figure}{0}
\setcounter{table}{0}

\renewcommand{\thetable}{\Alph{table}}
\renewcommand{\thefigure}{\Alph{figure}}

\section*{Acknowledgements}
The authors thank Supercomputing of Castile and Leon (SCAYLE. Leon, Spain) for assistance with the model training and GPU resources.

This work was supported by Spanish funds through Regional Funding Agency Institute for Business Competitiveness of Castile Leon (MACS.2 project “Investigación en tecnologías del ámbito de la movilidad autónoma, conectada, segura y sostenible”).

\section{Additional Implementation Details}
\label{sec:training}
Our implementation is based on PyTorch \cite{pytorch}. We train DarkIR (following LEDNet~\cite{lednet}) on the LOLBlur dataset. During training, we randomly crop $384\times 384$ patches, and apply standard flip and rotation augmentations. The mini-batch size is set to 32 using an H100 GPU. 

As our optimizer we use AdamW \cite{adamw} by setting $\beta_1=0.9$, $\beta_2=0.9$ and weight decay to $1e^{-3}$. The learning rate is initialized to $5e^{-4}$ and is updated by the cosine annealing strategy \cite{cosine} to a minimum of $1e^{-6}$.  We repeat this configuration for re-training the other methods in LOLBlur dataset. Note that we use the official open-source implementation of the other methods, or previously reported results.

The \textbf{multi-task} model was trained using the same setup. The only difference is the use of LOLv2 and LSRW as additional datasets. This model achieves essentially state-of-the-art results on real low-light enhancement benchmarks, while maintaining the performance on LOLBlur.

\section{Additional ablation Studies}

We studied the influence of the optimization losses. The results can be seen on Table \ref{tab:abla_loss}, where we can check that by introducing the $L_{edge}$, $L_{lol}$ and $L_{percep}$ the model achieves the best combination of distortion and perceptual metrics.

\begin{table}[t]
    \centering
    \caption{{Ablation study on our loss functions. We train DarkIR using different loss setups. Adding the perceptual loss ($L_{percep}$), edge loss ($L_{edge}$) and the architecture guiding loss ($L_{lol}$) helps to improve the overall performance.}}
    \setlength\tabcolsep{2pt}
    \resizebox{0.95\linewidth}{!}{
        \begin{tabular}{l c c c}
            \toprule
             & PSNR$\uparrow$ & SSIM$\uparrow$ & LPIPS$\downarrow$ \\
            \midrule
            $L_{pixel}$ & 26.34    & 0.856    & 0.205 \\ 
            \midrule
            $L_{pixel}$ + $L_{lol}$ & 26.19     & 0.861    & 0.197 \\ 
            \midrule
            $L_{pixel}$ + $L_{lol}$ + $L_{edge}$ & \underline{26.717}     & 0.874    & 0.182 \\ 
            \midrule
            $L_{pixel}$ + $L_{percep}$ + $L_{edge}$ & 26.61    & \textbf{0.877 }  & \textbf{0.171} \\ 
            \midrule
            $L_{pixel}$ + $L_{lol}$ + $L_{edge}$ + $L_{percep}$   & \textbf{26.9}    & \underline{0.874}    & \underline{0.176}   \\ 
            \bottomrule

        \end{tabular}
    }
    \label{tab:abla_loss}
\end{table}

\begin{figure}[t]
    \centering
    \includegraphics[width=\linewidth]{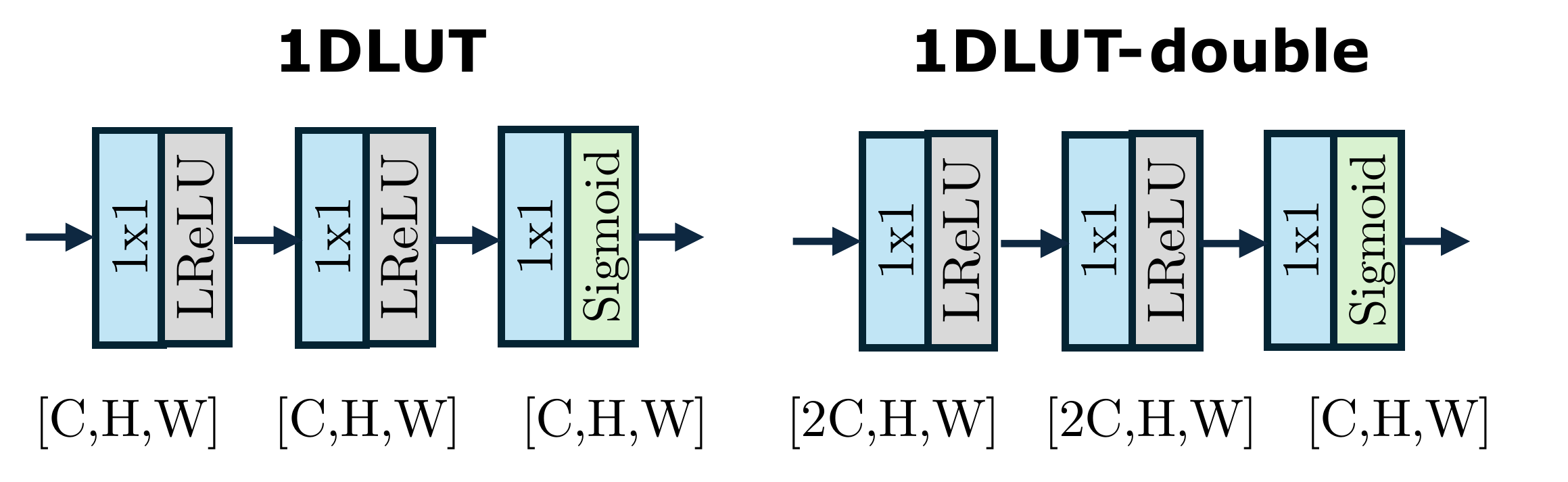}
    \caption{Neural blocks proposed for the feature propagation. The difference between both is the presence of a channel expansion in \textbf{1DLUT-double}.}
    \label{fig:skip}
\end{figure}

In addition, we studied different skip connections for the feature propagation between encoder and decoder. The proposed feature propagation takes the form of:

\begin{equation}
    y = f_{prop}(enc_{feat}) + dec_{feat}
\end{equation}

where $f_{prop}$ is the proposed feature block applied to the encoder features ($enc_{feat}$) added to the decoder features ($dec_{feat}$). Besides the CurveNLU proposed by LEDNet \cite{lednet} we evaluate the results obtained by using only depthwise convolutions in this given block. To sum up we incorporate a variation of this block that uses only point-wise convolutions, resembling the behaviour of a look-up table (LUT). Figure \ref{fig:skip} represents the proposed 1DLUT variations. In Table \ref{tab:abla_skip} the results of this ablation study are showcased. We see that the single addition, i.e $f_{prop}=Identity$ gets the best performance, so we did not consider adding any of the discussed blocks to the DarkIR architecture. 

\begin{table}[t]
    \centering
    \caption{Ablation study on the feature propagation between encoder and decoder. We found simple addition to be optimal.
    }
    \setlength\tabcolsep{2pt}
    \resizebox{1\linewidth}{!}{
        \begin{tabular}{l c c c c c}
            \toprule
                             & Params$\downarrow$ (M)   & MACs$\downarrow$ (G)     & PSNR$\uparrow$ & SSIM$\uparrow$ & LPIPS$\downarrow$ \\\midrule
          CurveNLU & 4.05             & 14.14                  & 26.55     & 0.872    & 0.176 \\ \midrule
          CurveNLU-DepthWise & 3.33             & 7.42                  & 26.64     & 0.872    & 0.177 \\ \midrule              
          1DLUT & 3.39             & 7.87                  & 26.69     & 0.873    & \textbf{0.175} \\ \midrule
          1DLUT-double & 3.49             & 8.9                  & 26.63     & \textbf{0.875}    & \underline{0.175} \\ \midrule
          Single Addition   & \textbf{3.31}              & \textbf{7.25}                  & \textbf{26.9}     & \underline{0.874}    & 0.176   \\ \bottomrule

        \end{tabular}
    }
    \label{tab:abla_skip}
\end{table}    

\begin{table*}[t]
      \centering
        \renewcommand{\arraystretch}{1.2}
        \caption{Quantitative comparison on five \textbf{real-workd unpaired LLIE} datasets using the perceptual quality metrics BRISQUE~\cite{mittal2012nobrisque} and NIQE~\cite{mittal2012making}. We use reference results from \cite{cidnet2024}.}
        \resizebox{\linewidth}{!}{
        \begin{tabular}{c|cc|cc|cc|cc|cc}
        \toprule
        \textbf{LLIE}&
         \multicolumn{2}{c|}{\textbf{DICM}}& 
         \multicolumn{2}{c|}{\textbf{LIME}} & 
         \multicolumn{2}{c|}{\textbf{MEF}} & 
         \multicolumn{2}{c|}{\textbf{NPE}} & 
         \multicolumn{2}{c}{\textbf{VV}}\\

        ~ \textbf{Unpaired}&
            BRISQUE$\downarrow$&	NIQE$\downarrow$&
            BRISQUE$\downarrow$&	NIQE$\downarrow$&
            BRISQUE$\downarrow$&	NIQE$\downarrow$&
            BRISQUE$\downarrow$&	NIQE$\downarrow$&
            BRISQUE$\downarrow$&	NIQE$\downarrow$
            \\
            
            \midrule
            KinD~\cite{kind++}&
            48.72& 	5.15& 	
            39.91& 	5.03& 	
            49.94& 	5.47& 	
            36.85& 	4.98& 	
            50.56& 	4.30
\\
            ZeroDCE~\cite{Zero-DCE}&
            27.56& 	4.58 &
            \underline{20.44}& 	5.82 &
            17.32& 	4.93 &
            20.72& 	4.53 &
            34.66& 	4.81  
\\    
            RUAS~\cite{liu2021retinex}&
            38.75& 	5.21& 	
            27.59& 	4.26& 	
            23.68& 	3.83& 	
            47.85& 	5.53& 	
            38.37& 	4.29 
\\
            SNR-Net~\cite{snr_net}&
            37.35& 	4.71& 
            39.22& 	5.74& 	
            31.28& 	4.18& 	
            26.65& 	4.32& 	
            78.72& 	9.87
\\
             CIDNet~\cite{cidnet2024} &
            21.47&	3.79&	
            \textbf{16.25}& \underline{4.13}&
            \textbf{13.77}&	\underline{3.56}&
            \underline{18.92}&	\textbf{3.74}&	
            \underline{30.63}& \textbf{3.21}
\\
            \textbf{DarkIR-mt}&
            \textbf{18.69}& 	\textbf{3.76}& 	
            21.62& 	\textbf{4.07}& 	
            \underline{13.90}& 	\textbf{3.45}& 	
            \textbf{12.88}& 	\underline{3.99}& 	
            \textbf{26.87}& 	\underline{3.74} \\
            \bottomrule
        \end{tabular}
        }
        
        \label{tab:unpaired}
    \end{table*}%


\begin{figure*}[t]
    \centering
    \setlength{\tabcolsep}{1pt} 
    \begin{tabular}{c c c c c}

    \includegraphics[width=0.18\linewidth]{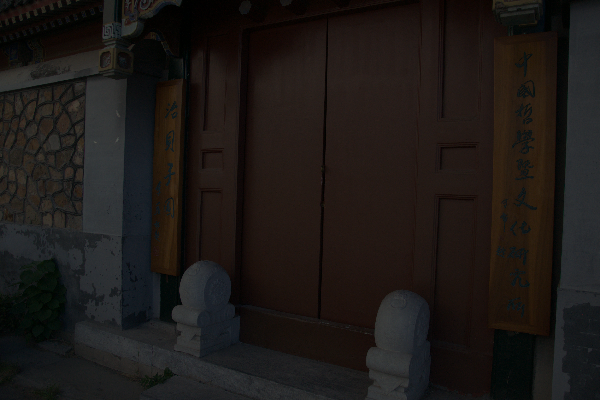} &

    \includegraphics[width=0.18\linewidth]{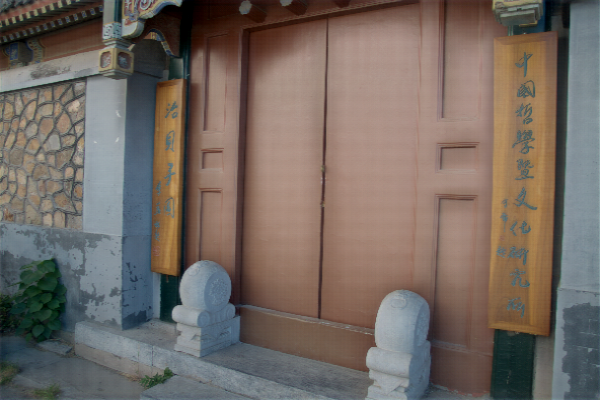} &

    \includegraphics[width=0.18\linewidth]{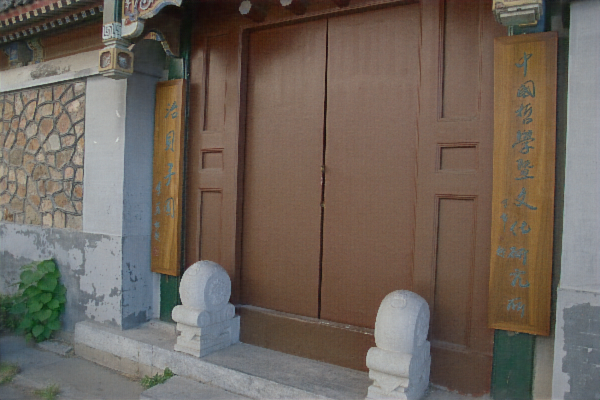} &

    \includegraphics[width=0.18\linewidth]{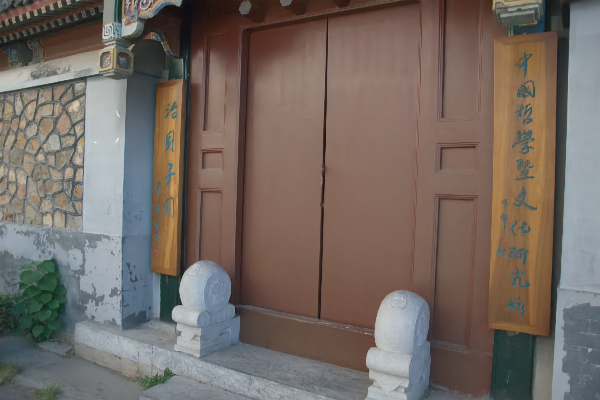} &

    \includegraphics[width=0.18\linewidth]{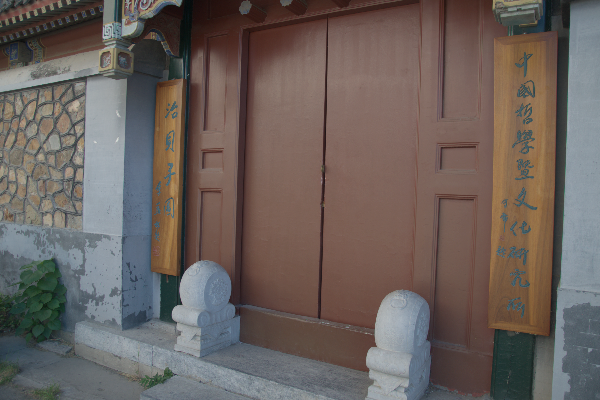} \\

    \includegraphics[width=0.18\linewidth]{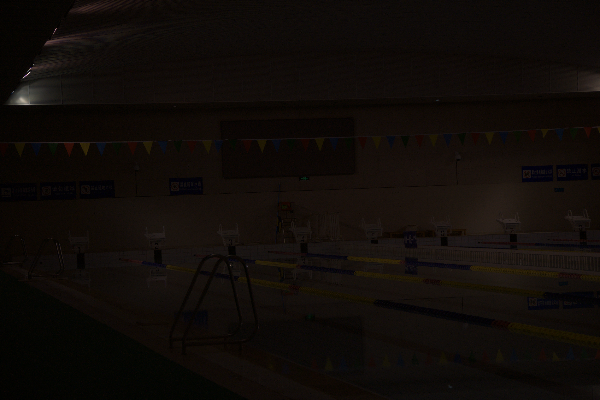} &

    \includegraphics[width=0.18\linewidth]{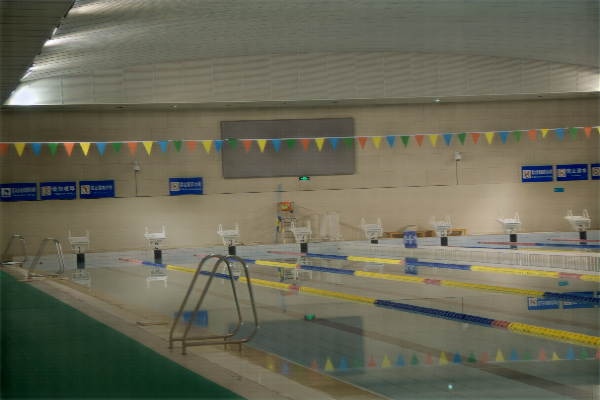} &

    \includegraphics[width=0.18\linewidth]{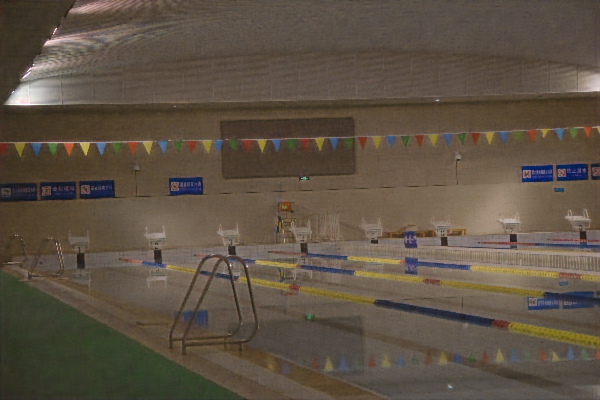} &

    \includegraphics[width=0.18\linewidth]{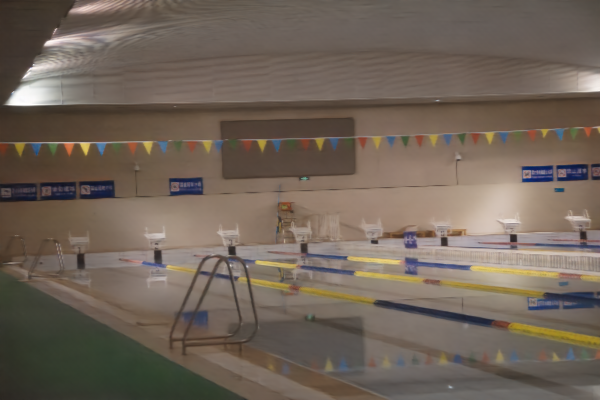} &

    \includegraphics[width=0.18\linewidth]{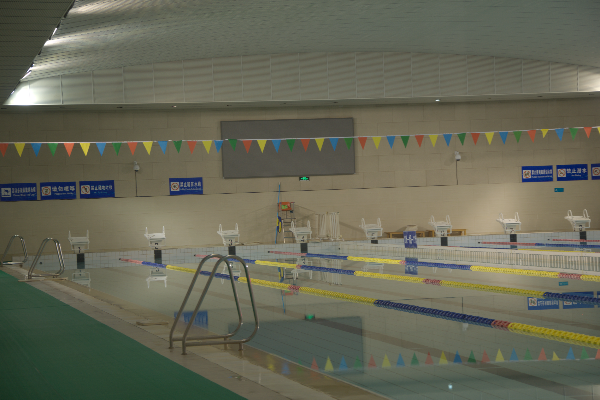} \\

    \includegraphics[width=0.18\linewidth]{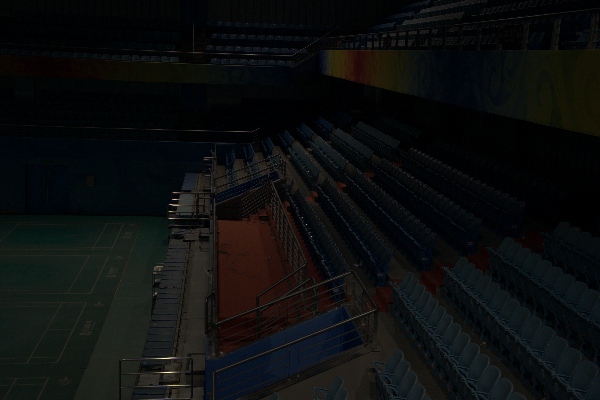} &

    \includegraphics[width=0.18\linewidth]{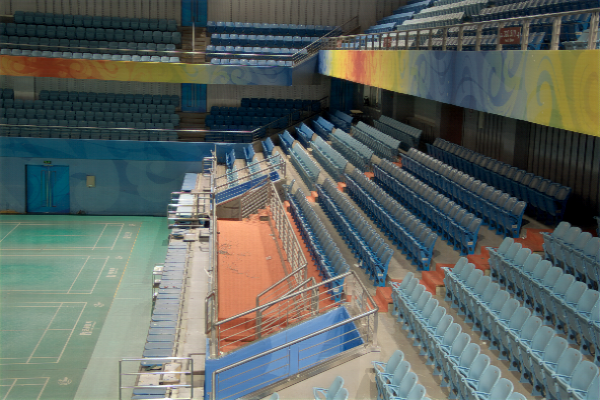} &

    \includegraphics[width=0.18\linewidth]{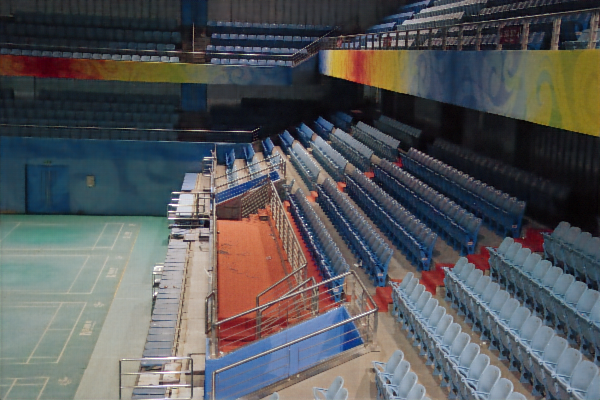} &

    \includegraphics[width=0.18\linewidth]{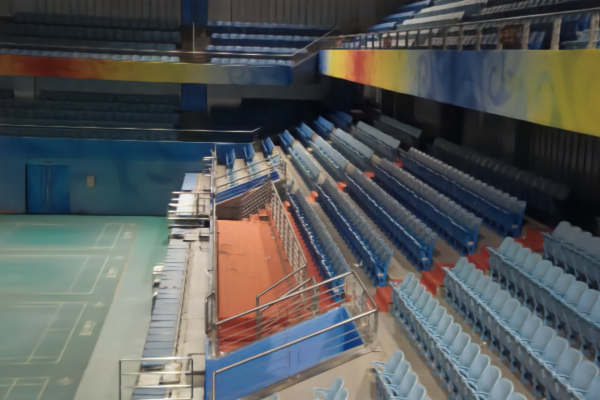} &

    \includegraphics[width=0.18\linewidth]{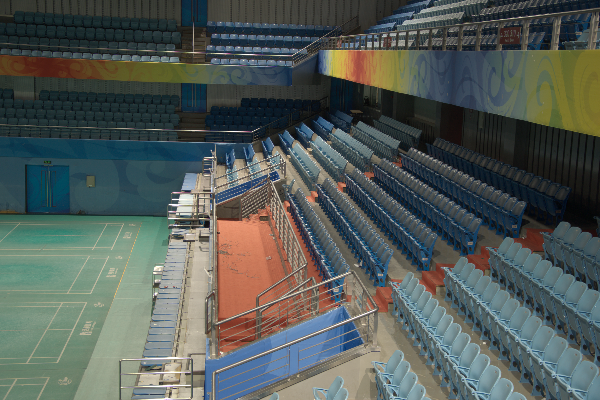} \\

    Input & SNR-Net & RetinexFormer & \textbf{DarkIR-m} & Ground Truth \\

    \end{tabular}
    \caption{Qualitative results compared with state of the art method RetinexFormer \cite{cai2023retinexformer} and SNR-Net \cite{snr_net} on \textbf{LOLv2-Real}.}
    \label{fig:v2real-supp}
\end{figure*}

\section{More quantitative results in unpaired data}
In addition to the results of unpaired Real-LOLBlur datasets, in Table \ref{tab:unpaired} we present the results of our model in 5 well-known unpaired datasets: LIME~\cite{guo2016lime}, DICM~\cite{lee2013contrastdicm}, MEF~\cite{ma2015perceptualmef}, NPE~\cite{wang2013naturalnessnpe} and VV~\cite{vonikakis2018evaluationvv}. We use the multi-task model trained LOLBLur and the LOL datasets. We report BRISQUE~\cite{mittal2012nobrisque} and NIQE~\cite{mittal2012making} metrics.

\section{More qualitative results in LLIE}

\begin{figure*}[t]
    \centering
    \setlength{\tabcolsep}{1pt} 
    \begin{tabular}{c c c c c}

    \includegraphics[width=0.18\linewidth]{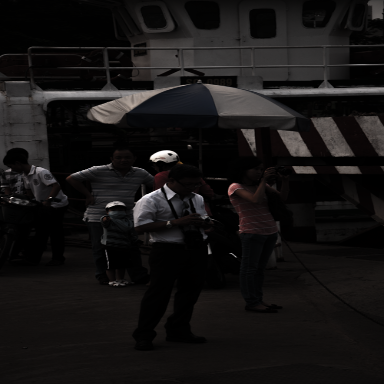} &

    \includegraphics[width=0.18\linewidth]{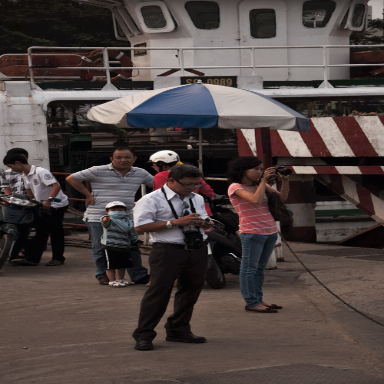} &

    \includegraphics[width=0.18\linewidth]{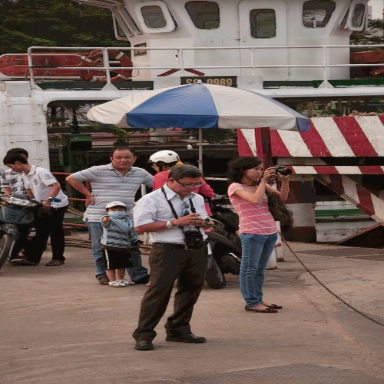} &

    \includegraphics[width=0.18\linewidth]{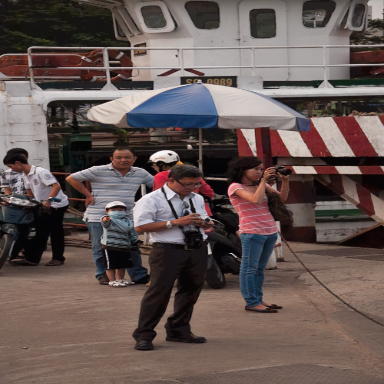} &

    \includegraphics[width=0.18\linewidth]{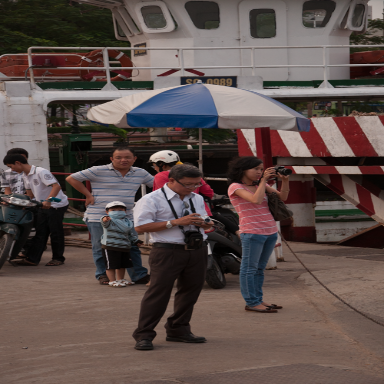} \\

    \includegraphics[width=0.18\linewidth]{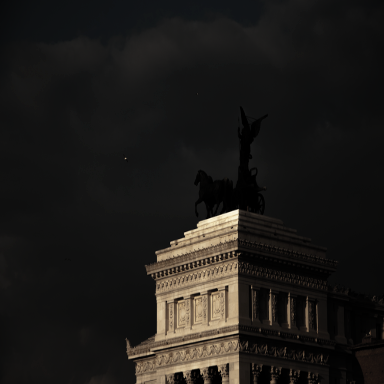} &

    \includegraphics[width=0.18\linewidth]{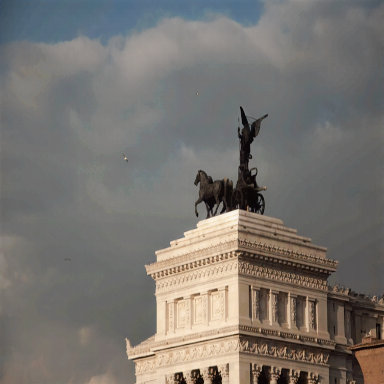} &

    \includegraphics[width=0.18\linewidth]{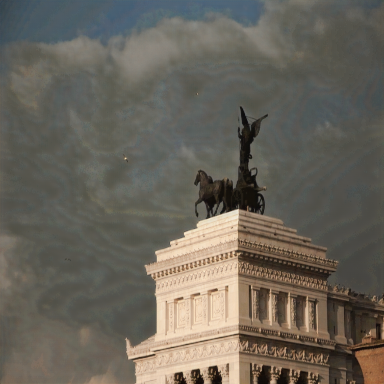} &

    \includegraphics[width=0.18\linewidth]{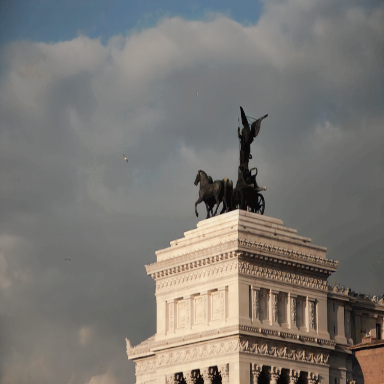} &

    \includegraphics[width=0.18\linewidth]{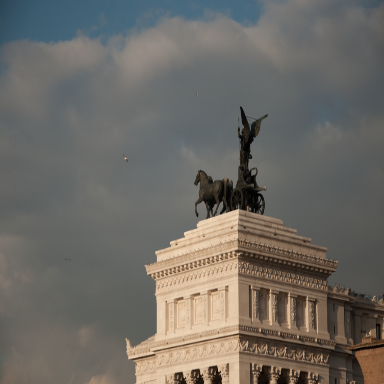} \\

    \includegraphics[width=0.18\linewidth]{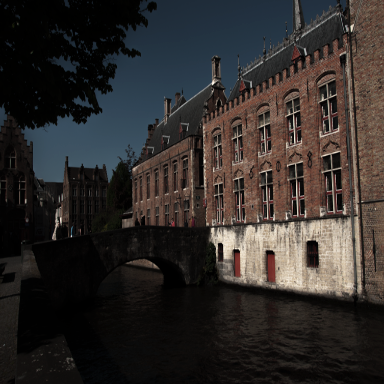} &

    \includegraphics[width=0.18\linewidth]{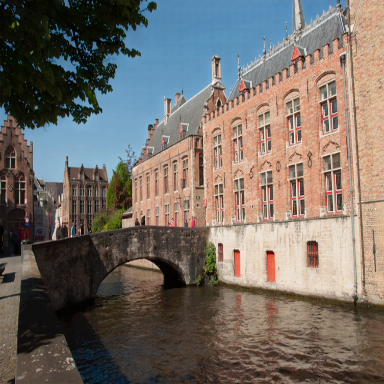} &

    \includegraphics[width=0.18\linewidth]{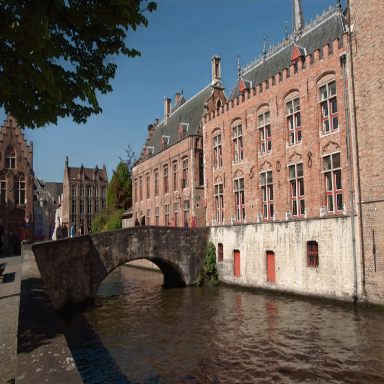} &

    \includegraphics[width=0.18\linewidth]{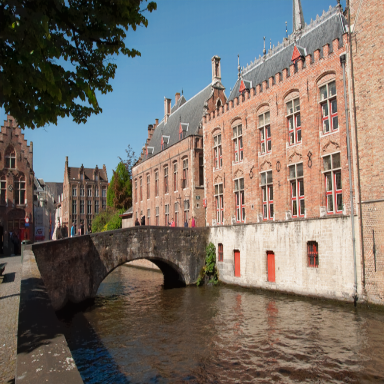} &

    \includegraphics[width=0.18\linewidth]{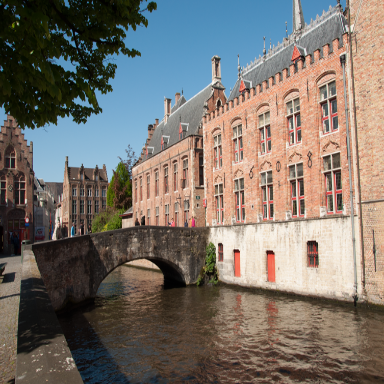} \\

    \includegraphics[width=0.18\linewidth]{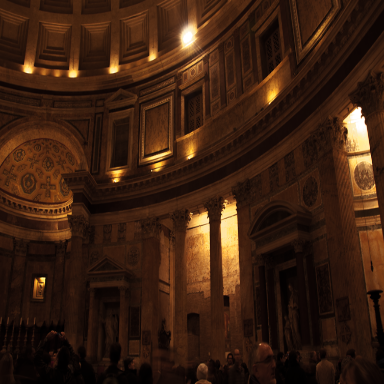} &

    \includegraphics[width=0.18\linewidth]{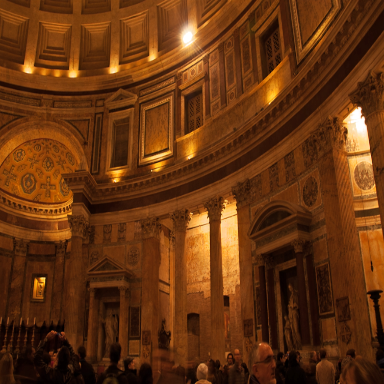} &

    \includegraphics[width=0.18\linewidth]{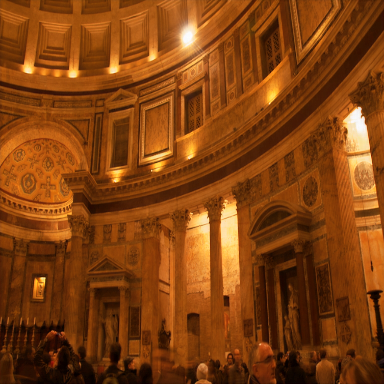} &

    \includegraphics[width=0.18\linewidth]{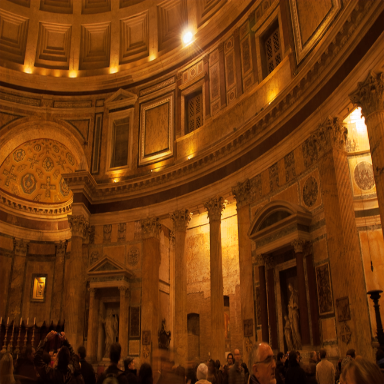} &

    \includegraphics[width=0.18\linewidth]{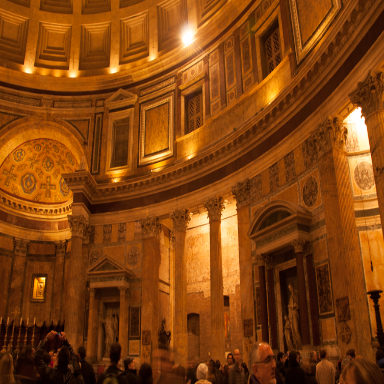} \\

    Input & SNR-Net & RetinexFormer & \textbf{DarkIR-m} & Ground Truth \\

    \end{tabular}
    \caption{Qualitative results compared with state of the art method RetinexFormer \cite{cai2023retinexformer} and SNR-Net \cite{snr_net} on \textbf{LOLv2-Synthetic}.}
    \label{fig:v2synth-supp}
\end{figure*}

\begin{figure*}[!ht]
    \centering
    \setlength{\tabcolsep}{1pt} 
    \resizebox{0.98\linewidth}{!}{
    \begin{tabular}{c c c c c c}

    \includegraphics[width=0.165\linewidth]{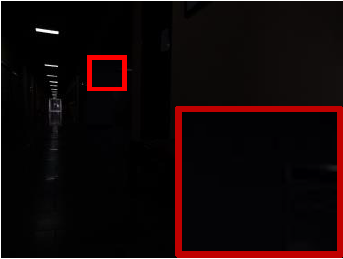} &

    \includegraphics[width=0.165\linewidth]{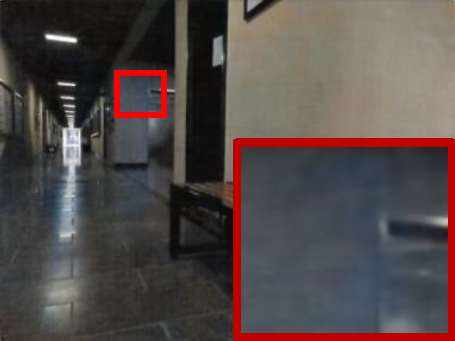} &

    \includegraphics[width=0.165\linewidth]{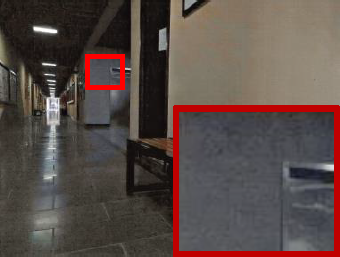} &

    \includegraphics[width=0.165\linewidth]{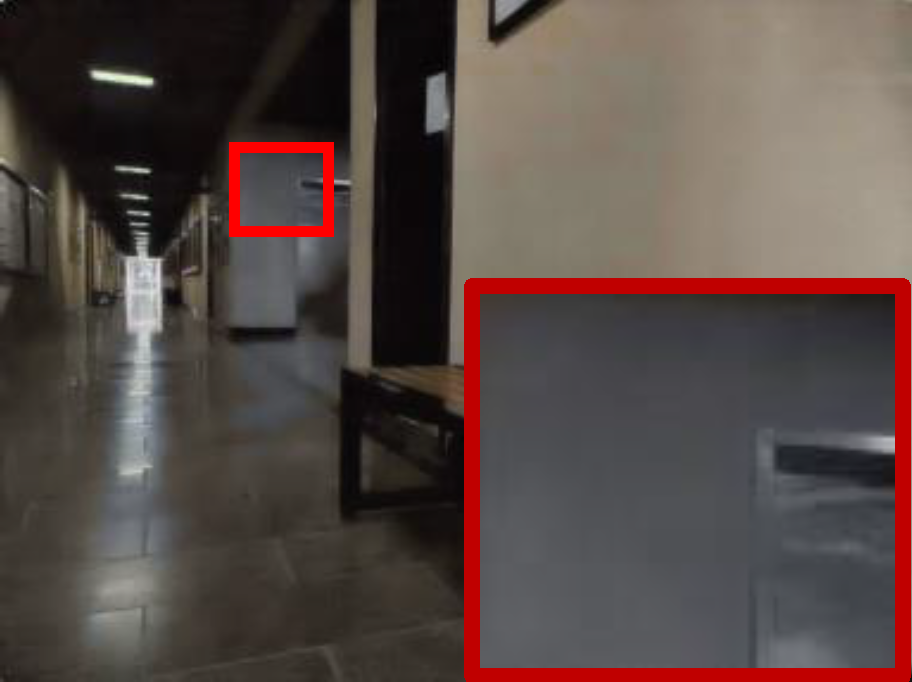} &

    \includegraphics[width=0.165\linewidth]{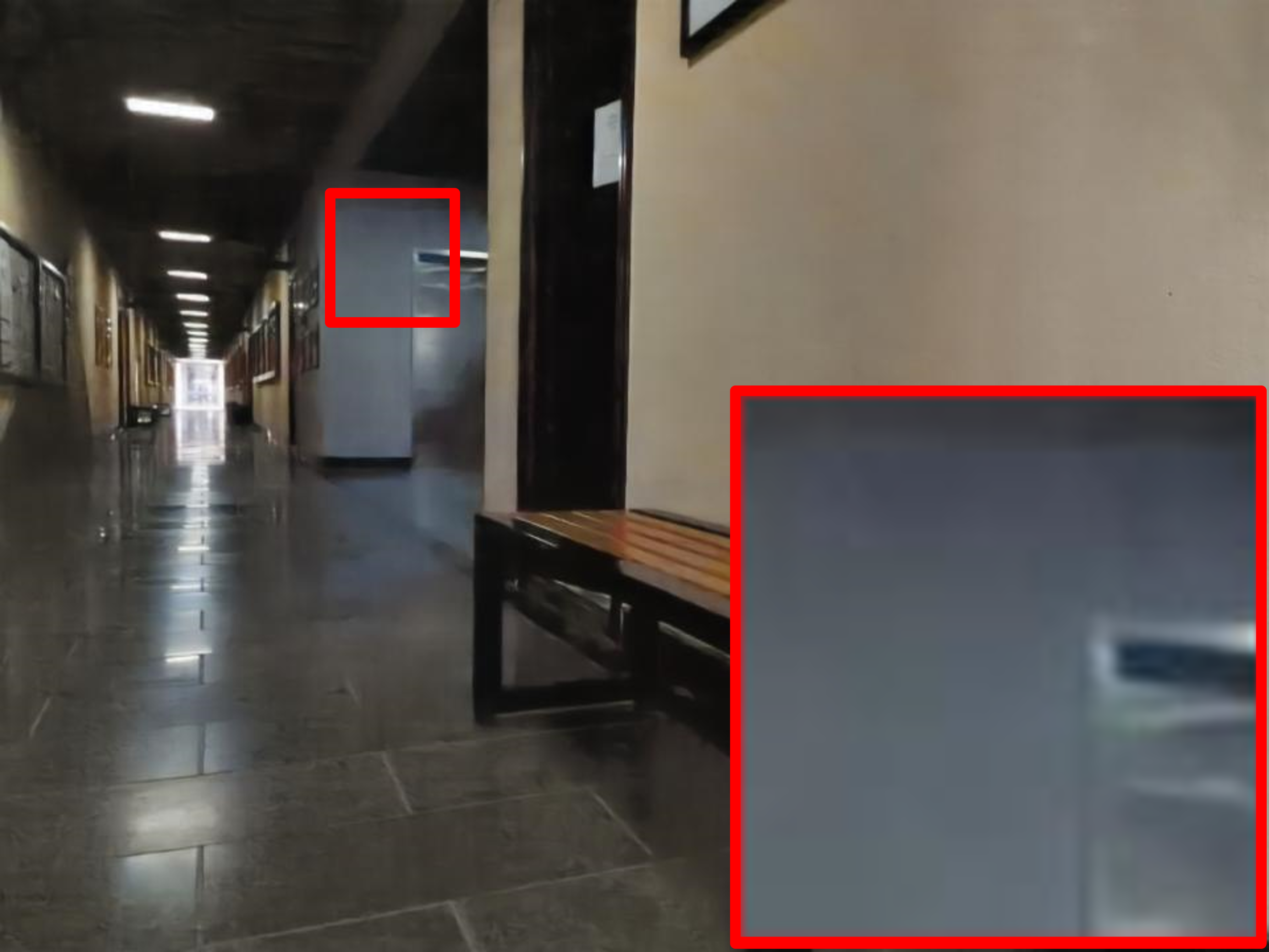} &

    \includegraphics[width=0.165\linewidth]{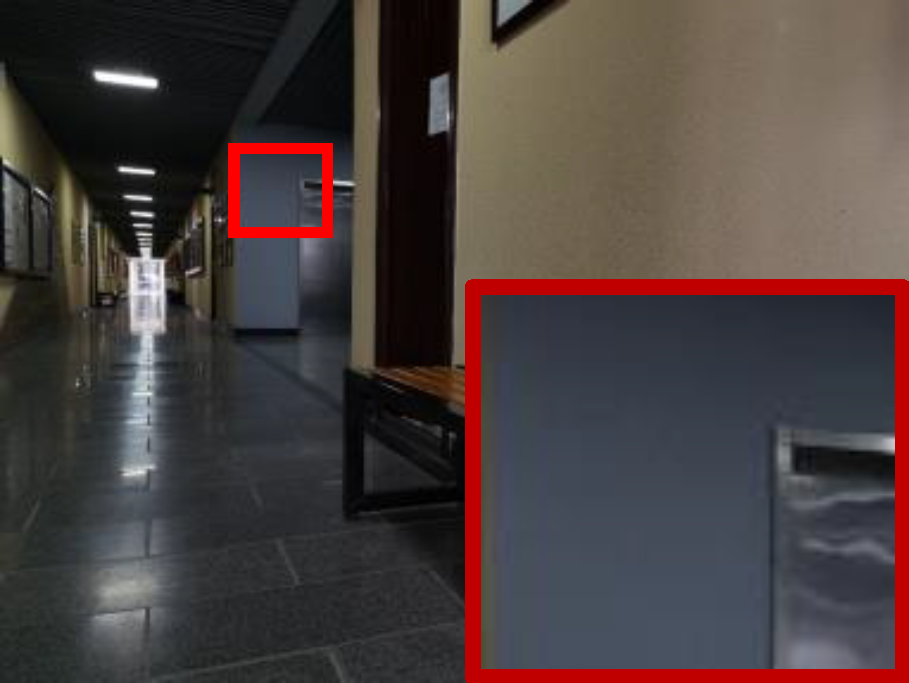} \\

    Input & DECNet & SNR-Net & FourLLIE & \bf{DarkIR-m} & Ground Truth \\

    \includegraphics[width=0.165\linewidth]{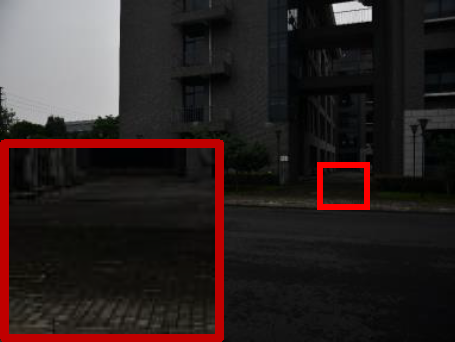} &

    \includegraphics[width=0.165\linewidth]{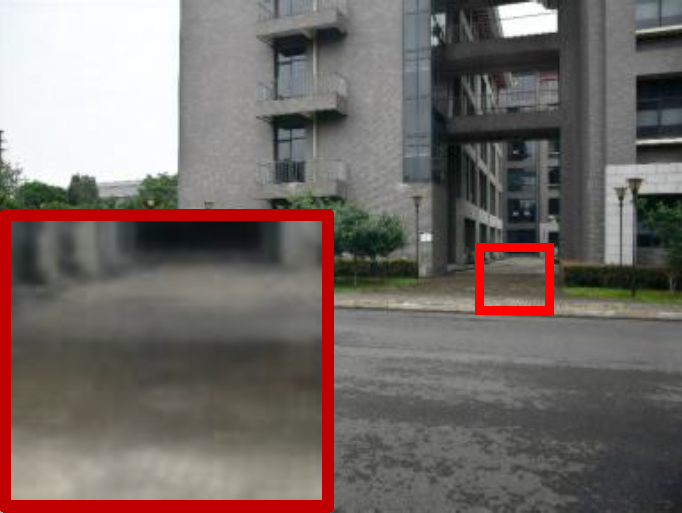} &

    \includegraphics[width=0.165\linewidth]{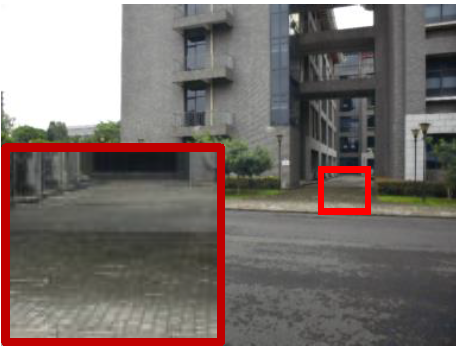} &

    \includegraphics[width=0.165\linewidth]{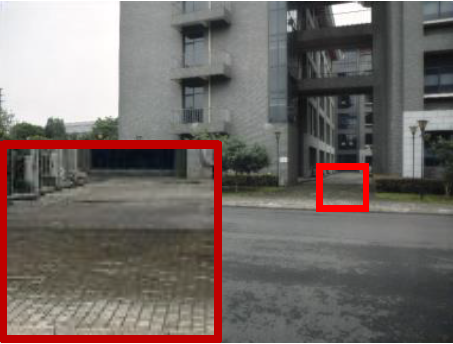} &

    \includegraphics[width=0.165\linewidth]{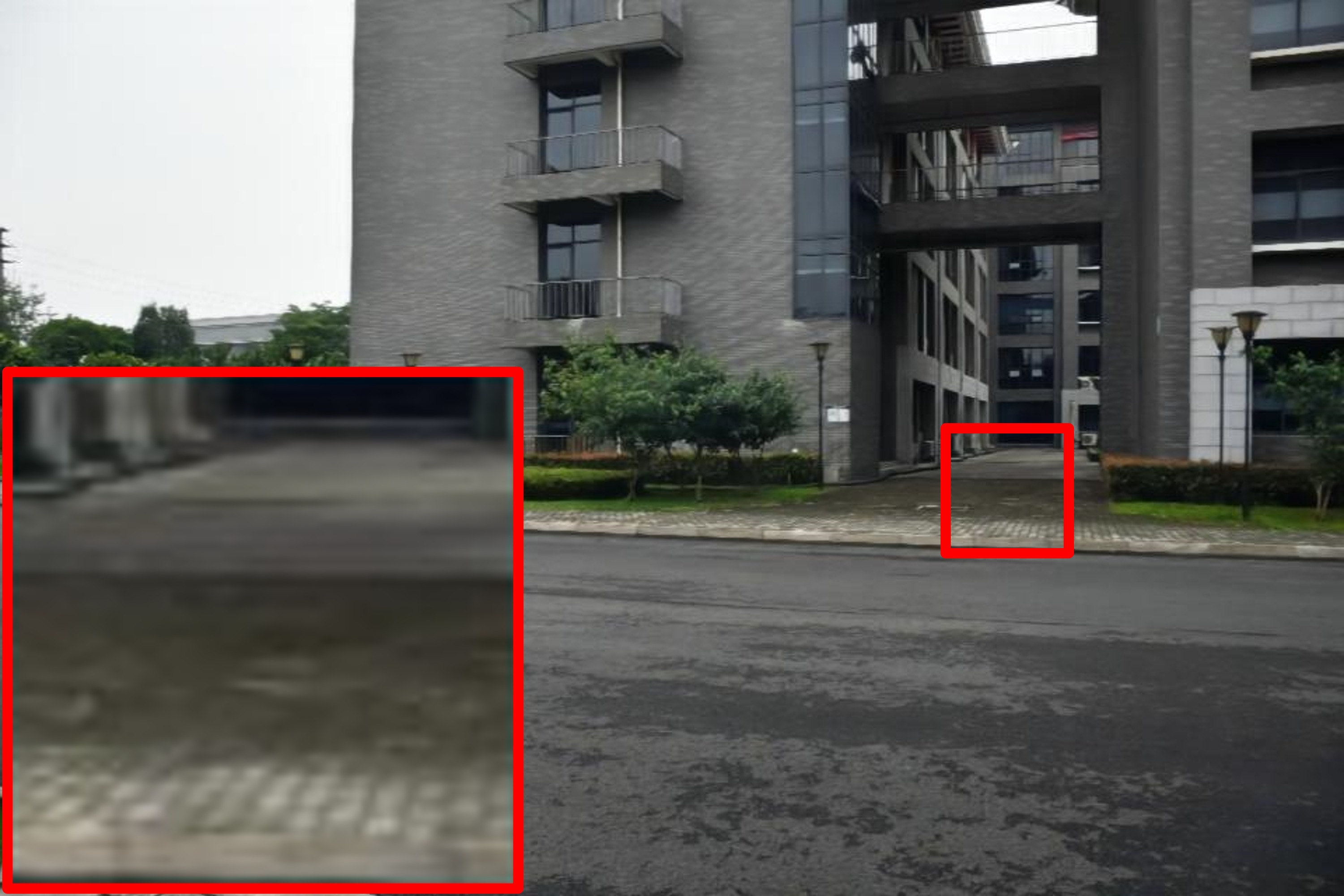} &

    \includegraphics[width=0.165\linewidth]{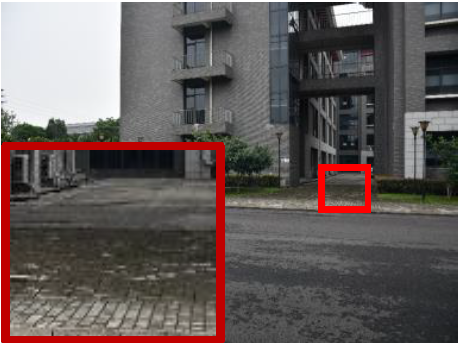} \\

    Input & MIRNet & SNR-Net & FourLLIE & \bf{DarkIR-m} & Ground Truth \\

    \end{tabular}
    }
    \caption{Qualitative results on the real-world dataset \textbf{LSRW-Huawei}~\cite{hai2023r2rnet} (top row) and \textbf{LSRW-Nikon}~\cite{hai2023r2rnet} (bottom row).}
    \label{fig:lsrw-supp}
\end{figure*}

\begin{figure*}[!ht]
    \centering
    \includegraphics[width=\linewidth]{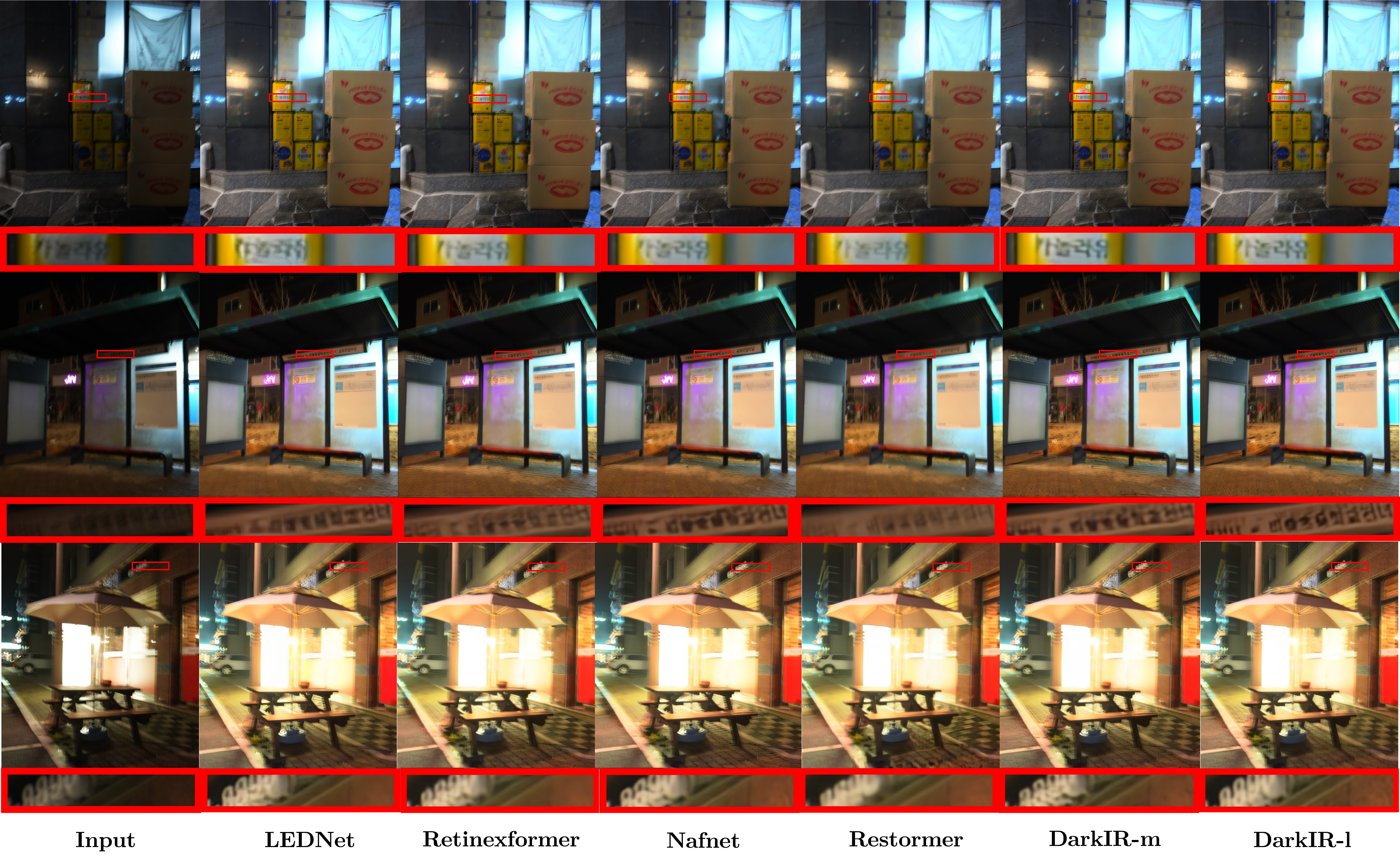}
    \caption{Qualitative results on \textbf{RealBlur-Night} \cite{realblur} images. (Zoom in for best view).}
    \label{fig:supp_realblur_night}
\end{figure*}

\begin{figure*}[!ht]
    \centering
    \setlength{\tabcolsep}{1pt} 
    \resizebox{0.8\linewidth}{!}{
    \begin{tabular}{c}
    \includegraphics[width=\linewidth]{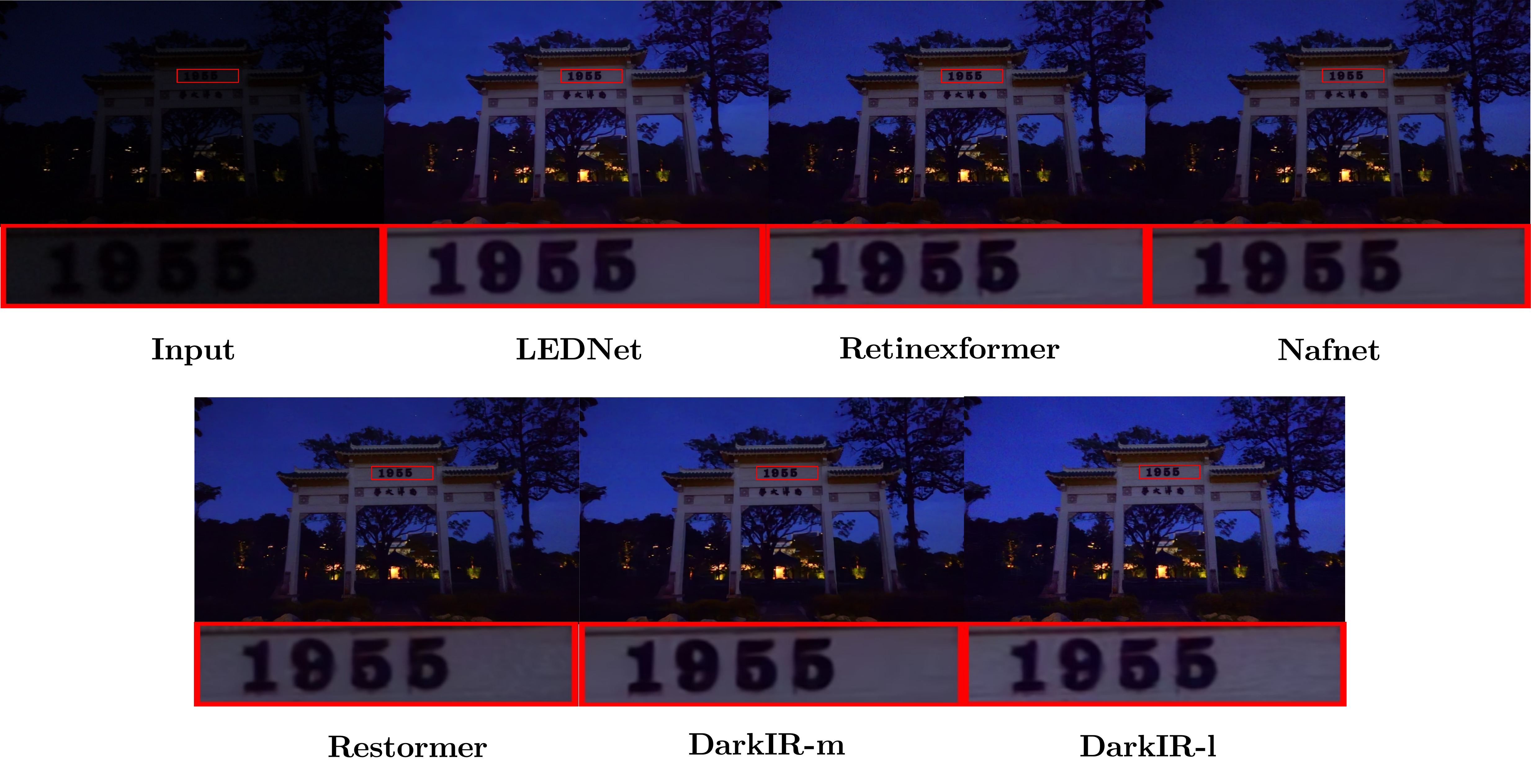} \\
    
    \includegraphics[width=\linewidth]{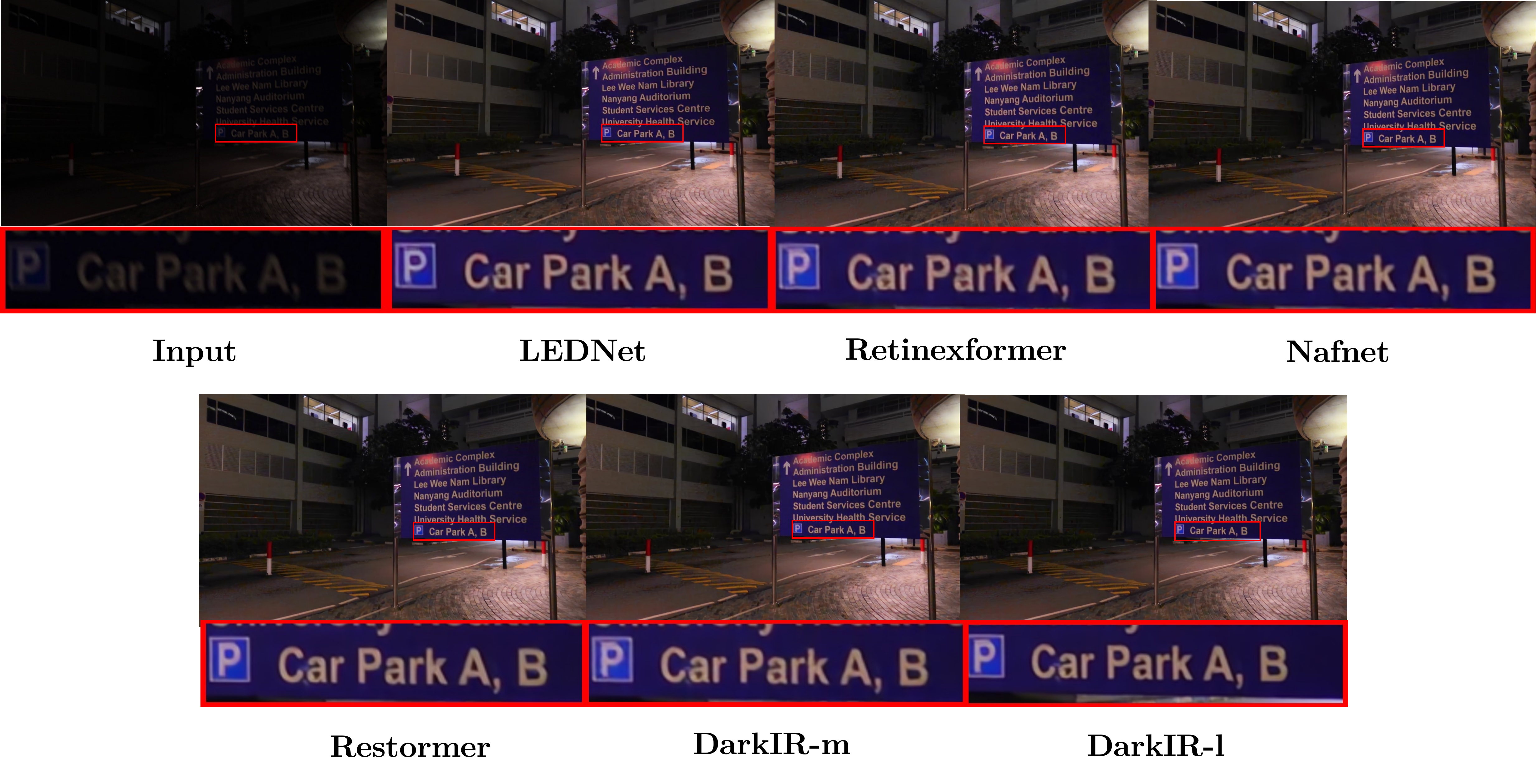} \\

    \includegraphics[width=\linewidth]{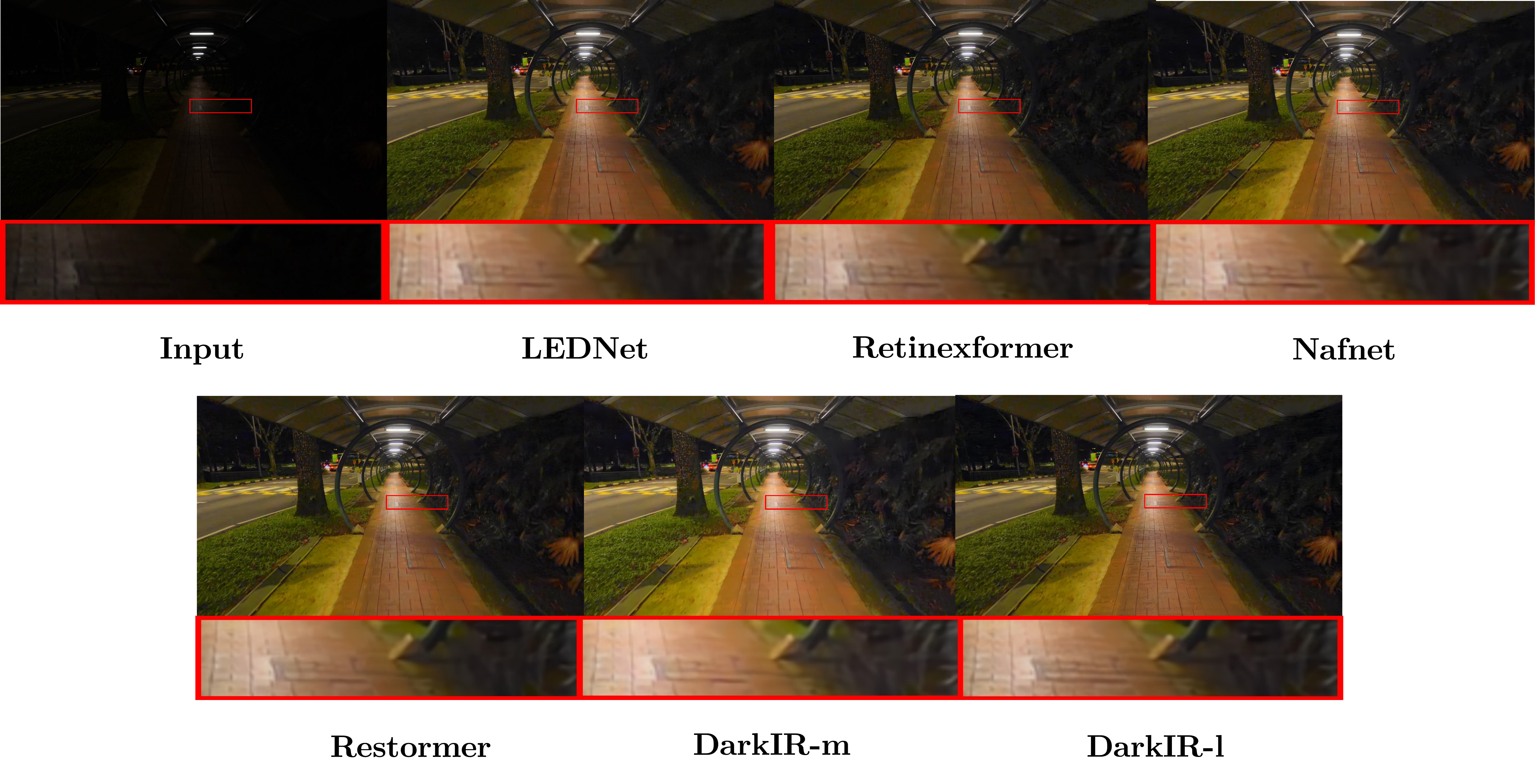} 

    \end{tabular}
    }
    \caption{Qualitative results in \textbf{Real-LOLBlur} \cite{lednet} dataset. (Zoom in for best view).}
    \label{fig:qualis_real_lolblur2}
\end{figure*}

\begin{figure*}[!ht]
    \centering
    \setlength{\tabcolsep}{1pt} 
    \resizebox{0.8\linewidth}{!}{
    \begin{tabular}{c}
    \includegraphics[width=\linewidth]{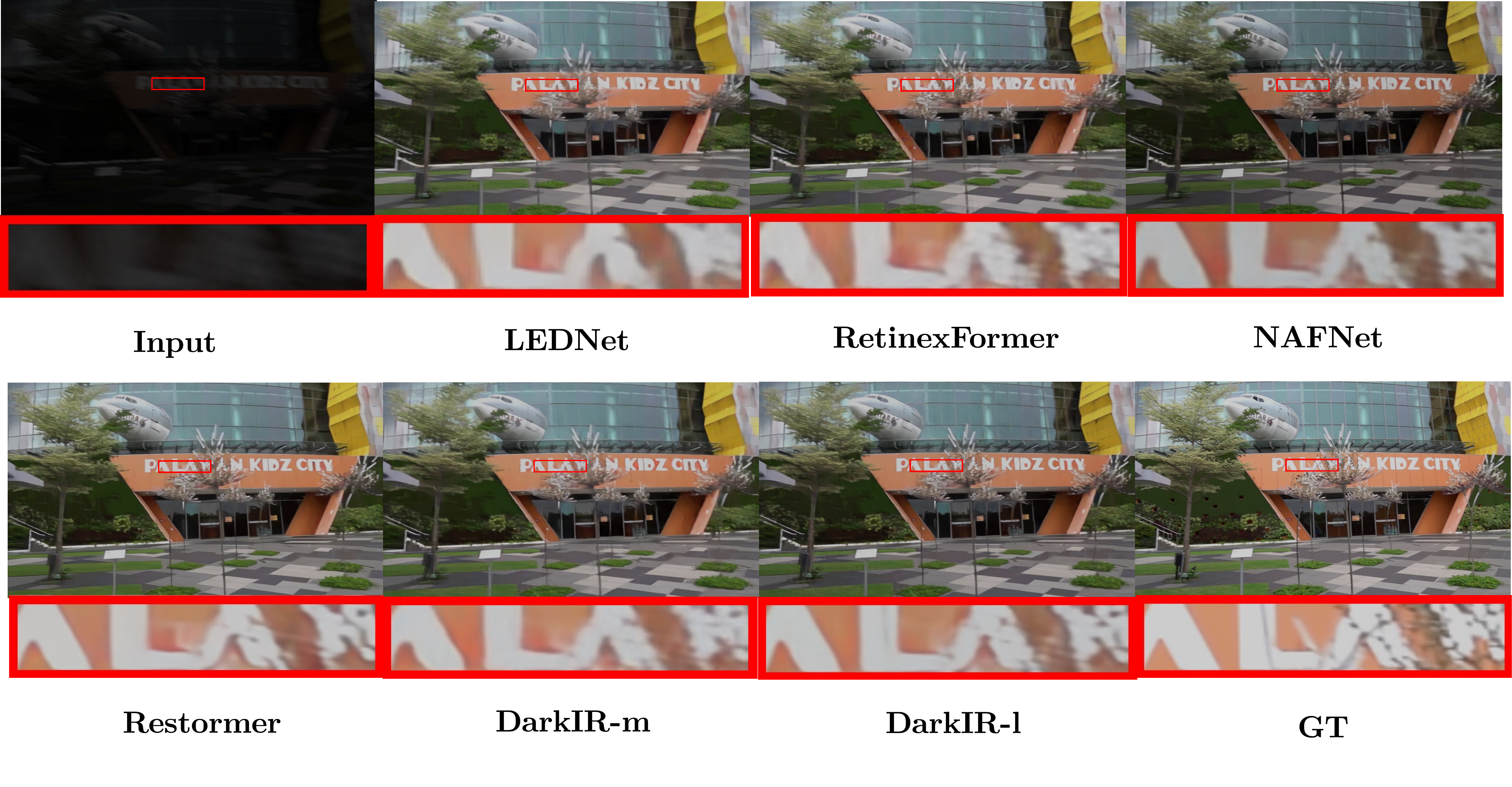} \\
    
    \includegraphics[width=\linewidth]{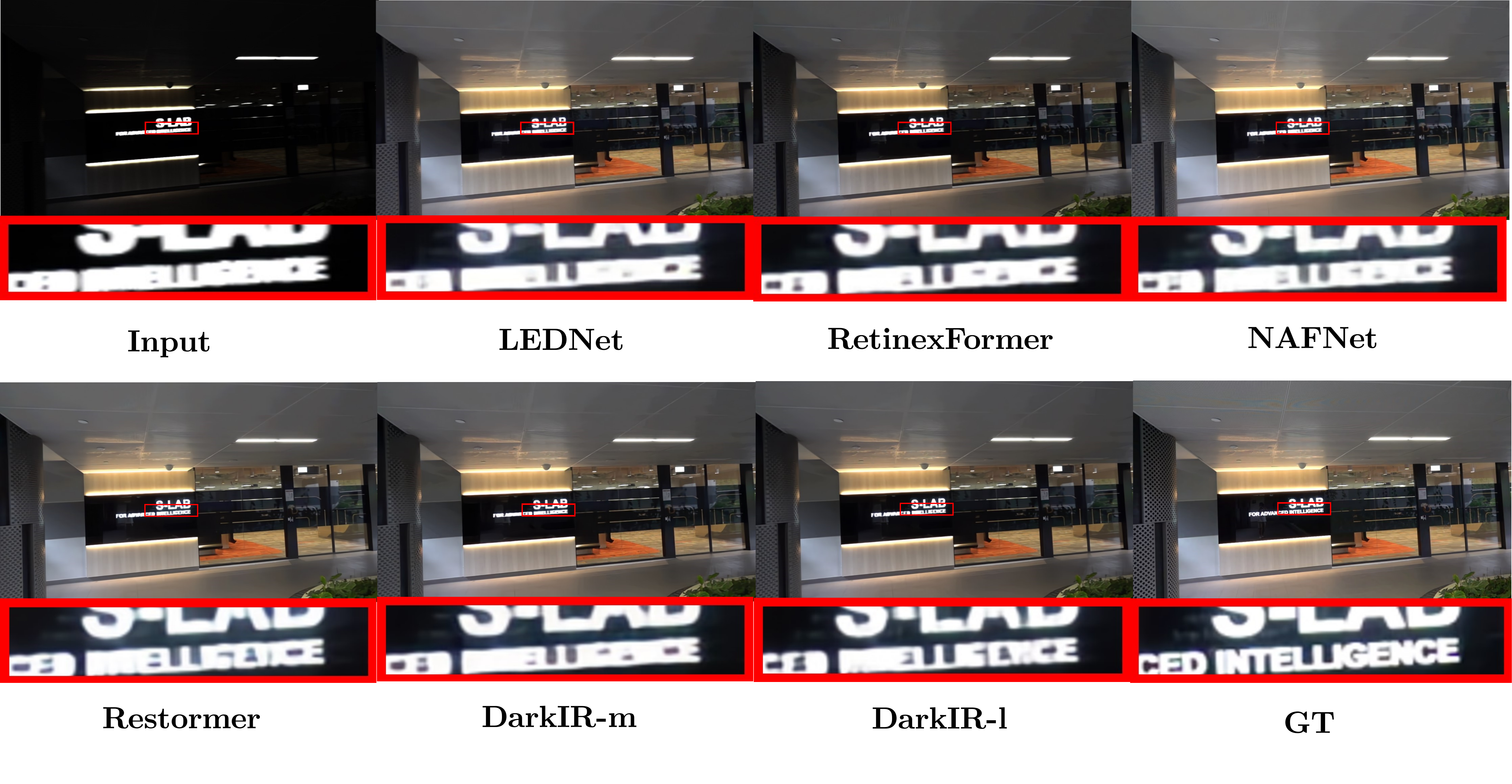} \\

    \includegraphics[width=\linewidth]{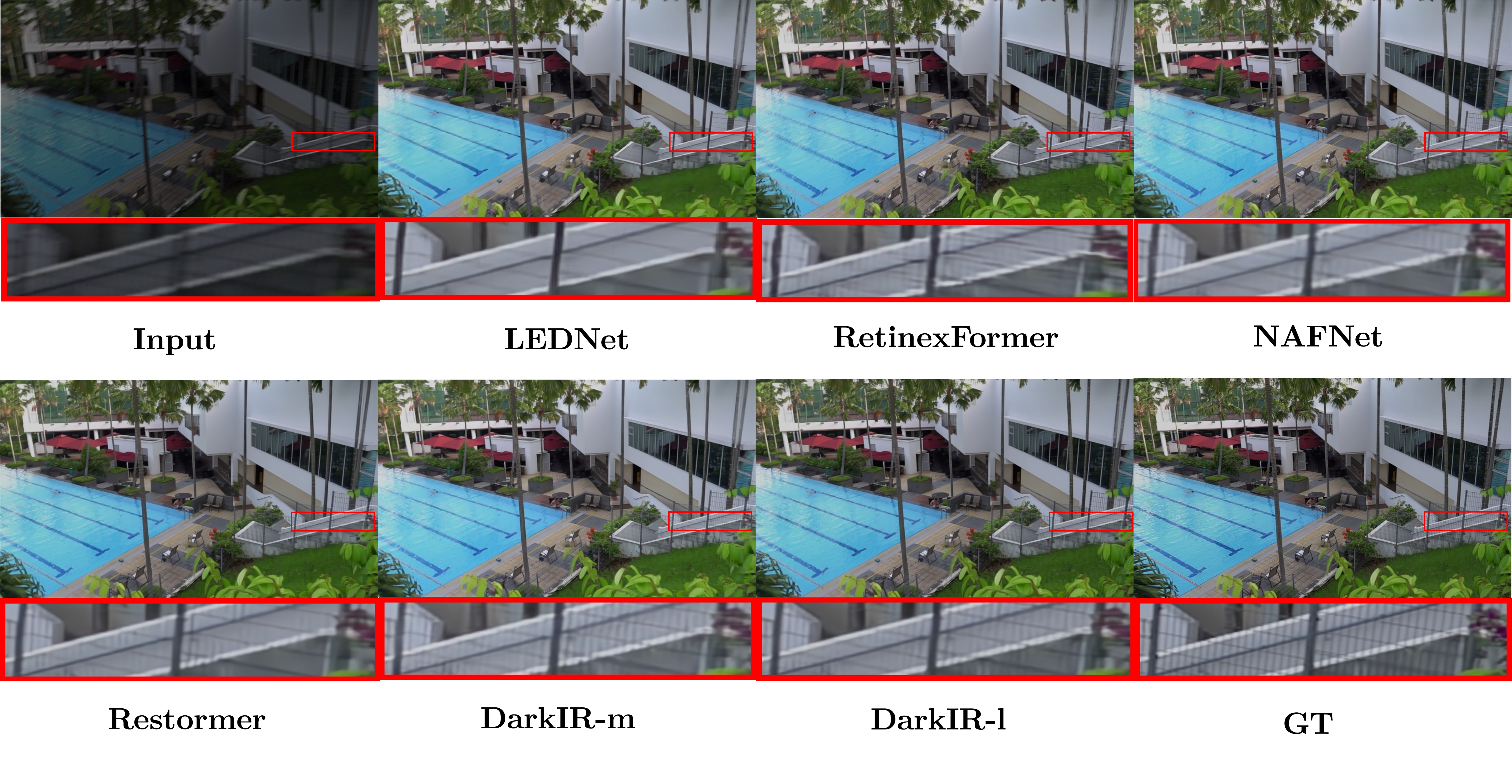} 

    \end{tabular}
    }
    \caption{Qualitative results in \textbf{LOLBlur} \cite{lednet} dataset. As we can see, \textbf{DarkIR} gets sharper and brighter results than the other methods. (Zoom in for best view).}
    \label{fig:qualis_lolblur2}
\end{figure*}

As we indicated in the paper, we present more qualitative results in Figures \ref{fig:v2real-supp}, \ref{fig:v2synth-supp} and \ref{fig:lsrw-supp}. These results showcase the power of our \underline{multi-task model} for LLIE restoration. Then, in Figures \ref{fig:supp_realblur_night}, \ref{fig:qualis_real_lolblur2} and \ref{fig:qualis_lolblur2} we show more qualitative results on the LLIE-Deblurring task.

